\documentclass[final]{cvpr}

\usepackage{times}
\usepackage{epsfig}
\usepackage{graphicx}
\usepackage{amsmath}
\usepackage{amssymb}
\usepackage{booktabs}
\usepackage{threeparttable}
\usepackage{subcaption}
\usepackage[misc]{ifsym}
\usepackage{lipsum}

\def\eg{\emph{e.g.}}

\def\ie{\emph{i.e.}}

\def\etal{\emph{et al.}}

\newcommand{\Tref}[1]{Table~\ref{#1}}

\newcommand{\Fref}[1]{Figure~\ref{#1}}
\newcommand{\Sref}[1]{Section~\ref{#1}}

\newcommand{\bG}{\mathbf{G}}
\newcommand{\bP}{\mathbf{P}}
\newcommand{\bH}{\mathbf{H}}

\newcommand{\bI}{\mathbf{I}}
\newcommand{\bA}{\mathbf{A}}
\newcommand{\bS}{\mathbf{S}}

\newcommand{\bN}{\mathbf{N}}

\newcommand{\edwardzhu}[1]{{{#1}}}

\usepackage[pagebackref=true,breaklinks=true,colorlinks,bookmarks=false]{hyperref}

\usepackage{tabularx}
\newcolumntype{L}[1]{>{\raggedright\arraybackslash}p{#1}}
\newcolumntype{C}[1]{>{\centering\arraybackslash}p{#1}}
\newcolumntype{R}[1]{>{\raggedleft\arraybackslash}p{#1}}

\def\cvprPaperID{3102} 
\def\confYear{CVPR 2021}
\begin{document}

\title{Spatially-Varying Outdoor Lighting Estimation from Intrinsics}



\author{Yongjie Zhu\textsuperscript{1 $\dagger$}\quad Yinda Zhang\textsuperscript{2}\quad Si Li\textsuperscript{1 $\ast$}\quad Boxin Shi\textsuperscript{3, 4 $\ast$}\\
  $^1$School of Artificial Intelligence, Beijing University of Posts and Telecommunications\\
  $^2$Google\quad $^3$NELVT, Department of Computer Science and Technology, Peking University\\
  $^4$Institute for Artificial Intelligence, Peking University
}

\maketitle
\newcommand\blfootnote[1]{%
\begingroup
\renewcommand\thefootnote{}\footnote{#1}%
\addtocounter{footnote}{-1}%
\endgroup
}

\blfootnote{$^\ast$ Corresponding authors.
  $^\dagger$ Part of this work was finished as a visiting student at Peking University.
This work was supported by National Natural Science Foundation of China under Grant No. 61872012, 62088102, National Key R\&D Program of China (2019YFF0302902), and Beijing Academy of Artificial Intelligence (BAAI).}

\begin{abstract}
  We present SOLID-Net, a neural network for spatially-varying outdoor lighting estimation from a single outdoor image for any 2D pixel location. Previous work has used a unified sky environment map
  to represent outdoor lighting. Instead, we generate spatially-varying local lighting environment maps by combining global sky environment map with warped image information according to geometric
  information estimated from intrinsics. As no outdoor dataset with image and local lighting ground truth is readily available, we introduce the SOLID-Img dataset with physically-based rendered images and their corresponding intrinsic and lighting information. We train a deep neural network to regress intrinsic cues with physically-based constraints and use them to conduct global and local lightings estimation. Experiments on both synthetic and real datasets show that SOLID-Net significantly outperforms previous methods. 
\end{abstract}

\section{Introduction}

Estimating outdoor lighting from a single image is one of the fundamental problems in computer vision. By providing outdoor scene properties from the physical aspect, it has huge impact on many applications, \eg, face/body relighting, scene understanding, augmented reality (AR), and so on. 
This task is rather challenging since images are formed by conflating lighting with complex surface reflectance distribution and object geometry. In the outdoor scenario, existing solutions usually employ low-dimensional parametric models such as the Ho\v{s}ek-Wilkie (HW) sky model~\cite{outdoorlighting_param} with four parameters to fit the sky illumination. 
The capacity of parametric models is not sufficient to represent the complex real-world illumination, and a recent non-parametric approach using an autoencoder to learn the sky illumination model from a large-scale sky panorama dataset and encoding the lighting information from a single limited Field-of-View (FOV) image shows more promising results~\cite{deepsky}. 
\begin{figure}
  \centering
    \includegraphics[width=1\columnwidth]{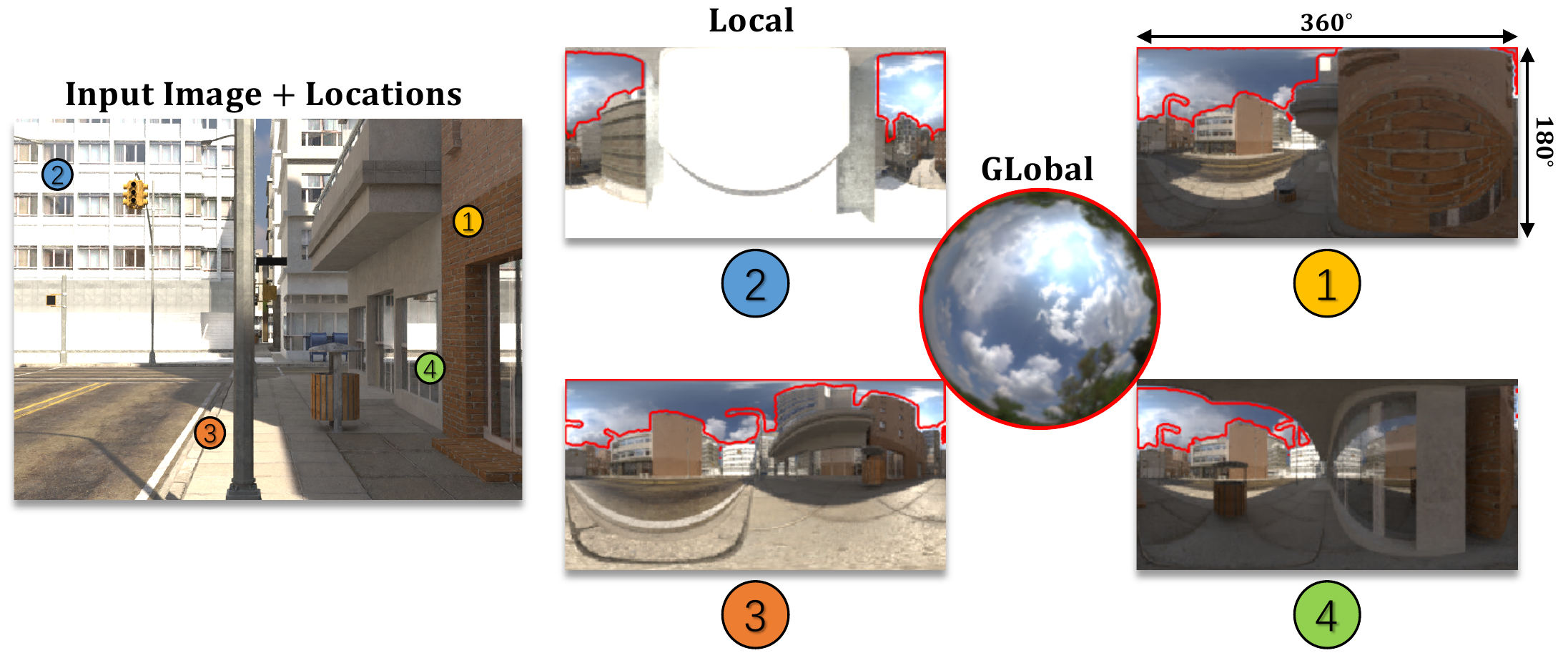}
    \caption{Given a single low-dynamic-range (LDR) image with limited FOV and a location in pixel coordinate (marked by numbers), SOLID-Net, for the first time, infers a panoramic HDR illumination map representing the light arriving from all directions at the location. Note that the global environment map (could be estimated using existing method~\cite{deepsky}) is only able to cover a small part of the local lighting (red contours).}
    \label{fig:teaser}
 \end{figure}

However, as far as we know, all existing outdoor lighting estimation methods~\cite{outdoorlighting_param,deepsky,allweather} only consider the outdoor illumination as a single global map without any spatially-varying consideration, \ie, the light probe is surrounded by an environment map that casts rays from infinitely far away. 
A spatially-varying lighting estimation has proved to be successful in indoor scenarios, which is achieved by modeling local indoor lighting using low-frequency parametric lighting represented by spherical harmonics (SH)~\cite{indoorlocallighting_SH,Barron2013} or panoramic environment map~\cite{indoorlocallighting_pano}. 

Extending spatially-varying lighting estimation from indoor to outdoor is non-trivial in three aspects: 1) The extremely high-dynamic-range (HDR) sunlight and the complicated sky light under different weather conditions make it more difficult to parameterize outdoor than indoor lighting~\cite{indoorlocallighting_SH,Barron2013}, while the existing non-parametric sky model~\cite{deepsky} treats it as a pure deep learning task without considering physics image formation constraint. 2) Non-parametric spatially-varying local lighting estimation is highly ill-posed, since different 3D locations should have different lighting observations and the majority of local observation is missing~\cite{indoorlocallighting_pano}. 3) HDR and panoramic images capturing local lighting and geometry information in outdoor are not yet available, despite there are many datasets for such a purpose in the indoor scenario by synthetically generating scenes from SUNCG~\cite{suncg} and Matterport3D~\cite{Matterport3D}. 

In this paper, we propose \textbf{SOLID-Net}, a neural network for \textbf{S}patially-varying \textbf{O}utdoor \textbf{L}ighting estimation using cues from \textbf{I}ntrinsic image \textbf{D}ecomposition, as shown in~\Fref{fig:teaser}. We tackle the three major challenges mentioned above by proposing a two-stage framework: 1) 
We train a single-in-multi-output CNN to decompose an input image into intrinsic parts: albedo (material-related), normal and plane distance (geometry-related), and shadow (lighting-related).
These intrinsics provide a physically-based shading constraint by fitting SH-represented global lighting with low-frequency information, which is then combined with extracted sky features from the input image to generate a non-parametric sky model like~\cite{deepsky}. 
2) With the estimated geometry from decomposed intrinsics, we further warp the input image and estimated shadow map with limited FOV to a spherical projection centered at the target location, which provides panoramic observation to reduce the ill-posed issue. This is then combined with global sky lighting from the previous step as input to train a multi-input-single-output CNN for complementing high-frequency local lighting estimation. 3) We use the Blender SceneCity~\cite{scenecity} to create city models that contain a large set of outdoor scenes and render a synthetic outdoor lighting estimation dataset with labeled location information and corresponding lighting effects using a physically-based path-tracer to facilitate the training of our network. SOLID-Net demonstrates significant improvements over other methods by making contributions in
\vspace{-3mm}
\begin{itemize}
  \setlength{\itemsep}{1pt}
  \setlength{\parskip}{0pt}
  \setlength{\parsep}{0pt}
  \item
  integrating shading constraint from intrinsic decomposition into the global sky lighting estimation;
  \item
  producing high-frequency local lighting estimation via panoramic warping and shadow map reference; and 
  \item
  building the first spatially-varying outdoor lighting estimation dataset with ground truth labels.
\end{itemize}

\begin{figure*}
  \centering
    \includegraphics[width=2\columnwidth]{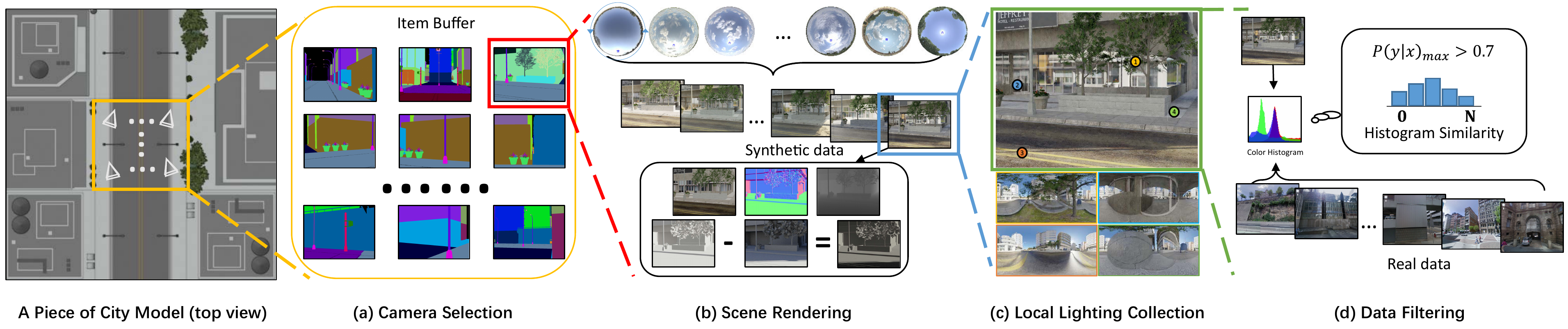}
    \caption{Pipeline of data generation and filtering for creating SOLID-Img dataset with physically based rendering.}
    \label{fig:dataset}
    \vspace{-2mm}
\end{figure*}

\vspace{-2mm}
\section{Related Work}
\vspace{2mm}
\noindent\textbf{Outdoor lighting estimation.}
Stumpfel~\etal~\cite{paul04} proposed to explicitly capture the HDR outdoor lighting environments that include the sun and sky with multiple exposures. Lalonde~\etal~\cite{lalonde12} first proposed lighting estimation from a single, generic outdoor scene. Their approach relied on multiple cues (such as shadows, shading, and sky appearance variation) extracted from the image. There are solutions using parametric models to represent outdoor lighting: Cheng~\etal~\cite{cheng2018} estimated lighting from the front and back camera of a mobile phone. However, they represented lighting using low-frequency SH, which does not appropriately model outdoor lighting. Hold-Geoffroy~\etal~\cite{outdoorlighting_param} learned to estimate Ho\v{s}ek-Wilkie (HW) sky model parameters from a single image, which is further extended by Zhang~\etal~\cite{allweather} with a more flexible parametric Lalonde-Matthews (LM) sky model. To include more information of the sky, Hold-Geoffroy~\etal~\cite{deepsky} designed an autoencoder to learn a non-parametric sky model from a large sky panorama dataset~\cite{skydatabase} and trained a network to learn the sky lighting from limited-FOV images.
LeGendre~\etal~\cite{legendre19} used a mobile phone camera with three different reflective spheres to capture lighting ground truth and used these data to train their deep model effectively, but these spheres are still global lighting probes. 

\vspace{2mm}
\noindent\textbf{Local lighting estimation.}
A direct way of obtaining the local lighting of an environment is to capture the lighting intensity at a target location using a probe of known shape. Debevec~\etal~\cite{paul98} showed that HDR environment maps can be captured with a reflective metallic sphere captuerd with the scene. Barron an Malik~\cite{Barron2013} decomposed the scene into intrinsic components including spatially-varying SH-based lighting, but it required an RGBD image as input and relied on hand-crafted priors. To learn local lighting representation, Garon~\etal~\cite{indoorlocallighting_SH} predicted fifth-order SH coefficients from an input image and local patches with synthetic data. In more recent progress, Li~\etal~\cite{indoorlighting_inverserender} proposed a dense spherical Gaussian lighting representation with differentiable rendering to conduct scene editing. But all the methods mentioned above only considered indoor parametric lighting and are hard to be extended to outdoor lighting. Song~\etal~\cite{indoorlocallighting_pano} proposed a cascaded model (denoted as NeurIllum for brevity) to recover high-frequency local lighting with warped color image according to recovered geometry, which showed promising texture details but the lighting positions are sometimes less accurate due to the loss of massive information in the panorama.  
\vspace{-5mm}
\section{Dataset}
\label{sec:dataset}
A large dataset containing HDR images and their corresponding illumination measured at different locations in a scene is required to learn to estimate outdoor intrinsics and local lightings.
Existing outdoor panorama datasets, such as~\cite{deepsky, ldr2hdr} only provide a single global illumination map assuming distant lighting, which cannot be used to learn local lighting estimation.
To provide training data for solving ``SOLID" problem, we introduce the \textbf{SOLID-Img}, a dataset for \textbf{S}patially-varying \textbf{O}utdoor \textbf{L}ighting estimation with ground truth \textbf{I}ntrinsic \textbf{D}ecomposition labels and a large amount of rendered \textbf{Im}a\textbf{g}es as shown in~\Fref{fig:dataset}. 

\subsection{Data Generation}
We adopt 3D city models from the Blender SceneCity~\cite{scenecity} to create synthetic scenes. In Blender SceneCity, there are 450 unique objects in 80 material categories. The object models provide surface materials, including diffuse albedo, roughness, and transparency, which are used to obtain photo-realistic renderings.

\vspace{2mm}
\noindent\textbf{Camera setting.}
For each road block, we select a set of cameras with diverse views seeing most objects in the context, to provide comprehensive information for lighting estimation, as shown in~\Fref{fig:dataset}(a). Our process starts by selecting the ``best'' camera~\cite{physicallyrender} for each of the six horizontal view direction sectors in every road block. 
For each sector, we select the view with the highest percentage pixel coverage according to item buffer, as long as it has more than three object categories\footnote{More details are in the supplementary material.}.

\vspace{2mm}
\noindent\textbf{Scene rendering.}
We collect 70 HDR environment maps from HDRI Haven~\cite{hdrihaven} which cover different solar zenith angles from sunrise to sunset. To simulate different sunlight directions, we rotate each HDR environment map along the latitude direction with a random angle sampled uniformly in $[0^\circ, 60^\circ]$. 
Then we render images using the camera settings above and these HDR environment maps with the physically-based Blender Cycles rendering engine~\cite{blender}, to generate photo-realistic renderings. The resolution is set as $320 \times 240$ with a physically-based path tracer of 512 samples. We record the material buffer (diffuse albedo buffer, normal buffer, depth buffer) as intermediate ground truth. We represent 3D geometry using the surface normal and plane distance, and render both as suggested in~\cite{im2pano3d}. To render shadows, we set the whole scene as a single Lambertian material and render it twice with shadow turned on and off respectively, from which shadow maps are calculated by checking the difference, as shown in ~\Fref{fig:dataset}(b). 

\vspace{2mm}
\noindent\textbf{Local lighting collections.}
To obtain the ground truth of global lighting, we save the rotated environment maps with $256 \times 128$ solution.
To collect local lighting, we randomly sample 4 locations in the scene to render 4 local light probes. The image is split into 4 quadrants, and a random 2D coordinate is sampled uniformly in each quadrant (excluding the sky part and 5\% pixels near the image boundary). The 3D centers of local cameras are calculated by casting a ray from the camera recording the scene to the surface of the scene and getting the first intersection point. From that point, we move the local camera center $10$cm away along the plane surface normal to prevent large invalid pixels and render a local light probe at this position with $256 \times 128$ resolution. All local light probes are rendered in the equirectangular representation, as shown in ~\Fref{fig:dataset}(c). 

\subsection{Data Filtering}
Inspired by \cite{physicallyrender}, we remove low-quality renderings that are with different color distribution with natural images, \eg, with overly low or high intensities.
To obtain a prior color distribution on real images, we compute normalized color histogram for 1100 selected real images from the Google Street View Dataset~\cite{googlestreetview}.
For each rendered image, we calculate the histogram similarity with those from Google Street View as the sum of minimal value of each bin; and then we assign it with a score calculated as the largest histogram similarity by comparing it to all real images; finally, we select all images with color similarity score larger than $0.7$, as shown in~\Fref{fig:dataset}(d). This process selects 38000 images from the initially rendered image set, composing the SOILD-Img dataset. Then care is taken to split the dataset into the train/test set according to different lighting conditions.
\begin{figure*}
  \centering
    \includegraphics[width=2.1\columnwidth]{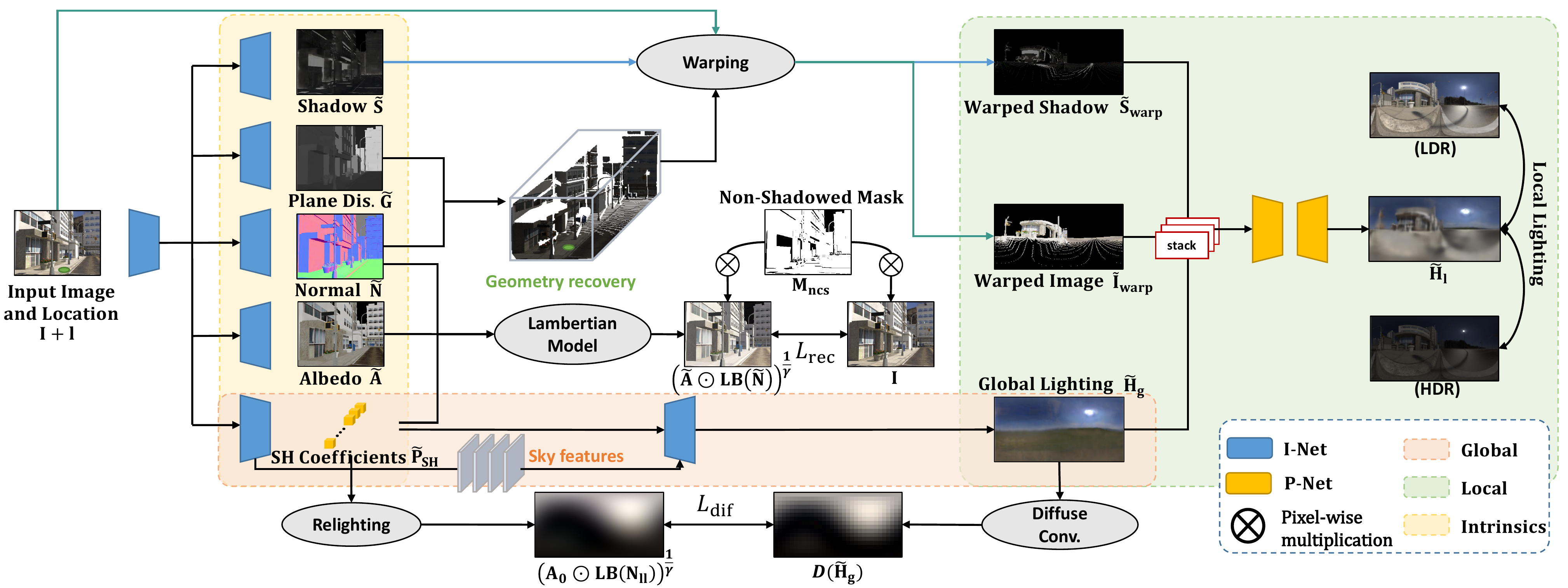}
    \caption{Pipeline of SOLID-Net. Stage 1: $\text{I-Net}$ takes input image $\mathbf{I}$ as input and estimate the intrinsic parts: $\tilde{\mathbf{A}}$, $\tilde{\mathbf{N}}$, $\tilde{\mathbf{G}}$, $\tilde{\mathbf{S}}$ and intermediate $\tilde{\mathbf{P}}_{\rm{SH}}$, then $\tilde{\mathbf{P}}_{\rm{SH}}$ is decoded with sky features to generate $\tilde{\mathbf{H}}_{\rm{g}}$. The recovered geometry from $\tilde{\mathbf{N}}$ and $\tilde{\mathbf{G}}$ is used to warp $\mathbf{I}$ and $\tilde{\mathbf{S}}$ into panoramic images $\tilde{\mathbf{I}}_{\rm{warp}}$ and $\tilde{\mathbf{S}}_{\rm{warp}}$ around an input location $\bold{l}$. Stage 2: $\text{P-Net}$ takes in warped images and $\tilde{\mathbf{H}}_{\rm{g}}$ to predict a high-frequency HDR spatially-varying lighting. 
    The whole network is trained in an end-to-end manner.}
    \label{fig:network}
    \vspace{-3mm}
\end{figure*}
\section{Method}
This section introduces the design methodology of SOLID-Net whose pipeline is shown in~\Fref{fig:network}. It is a two-stage framework that learns to reconstruct locally HDR outdoor environment maps, trained with the SOLID-Img dataset introduced in~\Sref{sec:dataset}.

\subsection{Problem Formulation}
We formulate illumination estimation as a regression problem. Given an LDR image $\bI$ with limited FOV and a selected pixel location ``$\bold{l}$'' in homogeneous coordinate $(x_0,y_0,1)^\top$, our model outputs an HDR illumination $\bH_{\rm{l}}$ centered around the 3D location of the pixel ``$\bold{l}$'' and a global sky environment HDR illumination $\bH_{\rm{g}}$, where both $\bH_{\rm{l}}$ and $\bH_{\rm{g}}$ are represented as a panoramic image with full FOV. 

\subsection{Network Architecture}
A straightforward approach to estimating the outdoor lighting from the scene would be to simply take the single limited FOV image as input, encode it into a feature map using a CNN, and feed the feature
map into a lighting-regression sub-network~\cite{lalonde17,deepsky}. Unsurprisingly, we find it results in outdoor lighting estimation with higher error (see~\Fref{fig:quan_sun_error}), presumably
because it is difficult for the network to understand how to extract full FOV lighting from a limited FOV image. One way to improve it is to bring regularization from the Lambertian rendering equation~\cite{inverserendernet}, which however is challenging for outdoor spatially-varying lighting estimation because: 1) Outdoor scenes have large areas of shadow occlusion which cannot be directly fitted by the Lambertian model. 2) SH lighting has a limited dynamic range and it is too smooth to represent sharp sky lighting and detailed texture. Therefore, we propose a two-stage framework to jointly solve these problems by: 1) proposing an intrinsic image decomposition network (denoted as \textbf{I-Net}) that takes a limited FOV image as input and estimates its intrinsic components as well as a global sky environment map and 2) designing a panoramic completion module (denoted as \textbf{P-Net}) that estimates local lighting from outputs in the previous stage and the input location.

\begin{figure*}
  \centering
    \includegraphics[width=2.1\columnwidth]{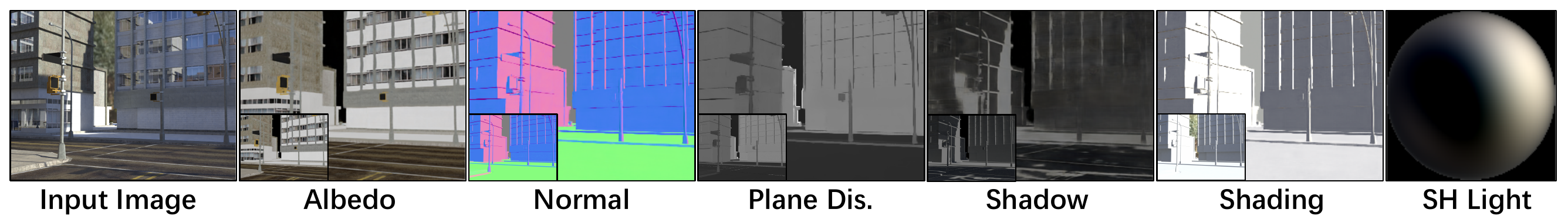}
    \caption{An example of intrinsic decomposition results using our SOLID-Img test dataset. Given an input image, our estimated albedo, normal, plane distance, shadow, and shading show close appearance to the ground truth (shown as insets).} 
    \label{fig:synthetic_results}
    \vspace{-2mm}
\end{figure*}
\begin{figure}
  \centering
    \includegraphics[width=1\columnwidth]{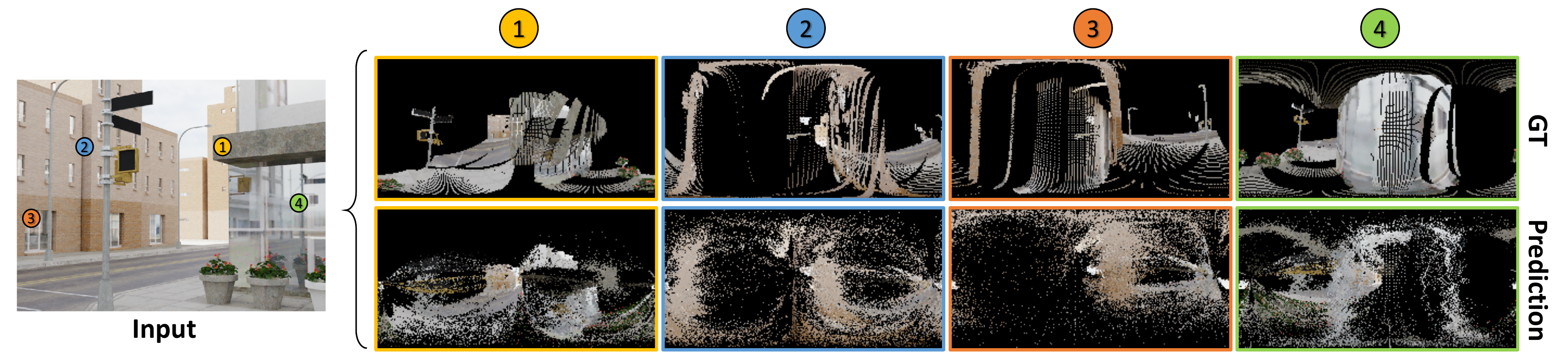}
    \caption{An example of panoramic warping. By using the estimated geometry-related intrinsics, we warp the observed image into panorama coordinates according to the input pixel location. (Please zoom-in for details.)}
    \label{fig:warping_results}
    \vspace{-2mm}
\end{figure}
  
\vspace{2mm}
\noindent\textbf{I-Net.}
As shown by blue blocks in Fig 3, I-Net takes a single limited-FOV LDR image $\bI$ as input and various outputs including diffuse albedo $\bA$, surface normal $\bN$, plane distant map $\bG$~\cite{im2pano3d}, shadow map $\bS$, and second-order SH coefficients $\bP_{\rm{SH}}$, and the sky environment map $\bH_{\rm{g}}$ generated from SH coefficients and sky features. We use a single encoder to capture global features of intrinsic information, and then use five decoders for $\bA$,  $\bN$, $\bG$, $\bS$, and $\bP_{\rm{SH}}$ followed by a decoder lighting branch for sky environment map regression. Skip links are used for preserving details. 
In particular the sky map regression, we use a fully-connected (FC) layer to process the output feature maps of lighting branch encoder to generate a latent vector of size 27 (second-order SH in RGB). For the decoder, we reshape this vector and upsample it 8 times, and then combine it by flipping padded sky features to generate a 256$\times$128 HDR sky environment map.
The lighting information encoded in $\bP_{\rm{SH}}$ can be considered as the low-frequency form of $\bH_{\rm{g}}$, and it is used to guide the recovery of the high-frequency sky environment maps with sky features extracted from the input images. In summary, I-Net predicts intrinsic components (examples are provided in~\Fref{fig:synthetic_results}) and sky environment map:
\begin{equation}
  \{\tilde{\mathbf{A}},\tilde{\mathbf{N}},\tilde{\mathbf{G}},\tilde{\mathbf{S}},\tilde{\mathbf{H}}_{\rm{g}}(\tilde{\mathbf{P}}_{\rm{SH}})\} =\text{I-Net}(\mathbf{I}).
  \label{equ:net1}
\end{equation}

\vspace{2mm}
\noindent\textbf{P-Net.}
The 3D location is calculated by the predicted normal vector ${\bold{n}}$ in $\tilde{\bN}$ and plane distance $p$ in $\tilde{\bG}$ by I-Net for each pixel. If the camera intrinsic matrix is fixed as $\bold{K}=[f_x,0,c_x;0,f_y,c_y;0,0,1]$ and 2D pixel locations $(x_i,y_i,1)^\top$ of the whole image are provided, we can reproject them as a 3D scene $\bold{P}=(x,y,z)^\top$ by $\bold{P}=-\frac{p}{{\bold{v}}\cdot{\bold{n}}}{\bold{v}}$, where ${\bold{v}}=(\frac{x_i-c_x}{f_x},\frac{y_i-c_y}{f_y},1)^\top$. 

By using $\bold{P}$ we warp the input image $\bI$  and estimated shadow map $\tilde{\mathbf{S}}$  and spatially align them with the output local lighting to provide panoramic observation.
First, we compute the camera location according to the input point position $\bold{l}$ and apply $10$cm translation (defined in~\Sref{sec:dataset}) along the normal direction of the supporting plane to align with training data. Second, we perform a panoramic warping through a forward projection using the estimated geometry and camera location to map pixels in $\bI$ and $\tilde{\mathbf{S}}$ as panoramic images (an example is provided in~\Fref{fig:warping_results}). The Z-buffer is computed to discard invisible points and the points without projected positions are set to 0. 

Since the local lightings share the same camera rotation, the sky parts in local lighting should be consistent, this motivates us to take the sky as an input to $\text{P-Net}$. 
As shown by orange blocks in Fig 3, P-Net concatenates the two incomplete panoramic image $\tilde{\bI}_{\rm{warp}}$ and $\tilde{\bS}_{\rm{warp}}$ and the global lighting estimated by I-Net $\tilde{\bH}_{\rm{g}}$ as inputs and outputs a dense pixel-wise prediction of local lighting panorama with full FOV and high-frequency details, as 
\vspace{-1mm}
\begin{equation}
  \vspace{-1mm}
  \tilde{\mathbf{H}}_{\rm{l}} = \text{P-Net}(\tilde{\mathbf{I}}_{\rm{warp}}(\bold{l}),\tilde{\mathbf{S}}_{\rm{warp}}(\bold{l}),\tilde{\mathbf{H}}_{\rm{g}}).
  \label{equ:net2}
\end{equation}
$\text{P-Net}$ is implemented as a fully convolutional U-Net~\cite{unet}. 

\subsection{Loss Functions}
\vspace{2mm}
\noindent\textbf{Direct supervision loss.}
Direct supervision $\mathcal{L}_{\rm{sup}1}$ for $\text{I-Net}$ is provided to 1) diffuse albedo predictions via $\rm{L}_2$ loss, 2) shadow predictions via $\rm{L}_2$ loss, 3) surface normal predictions via cosine loss, 4) plane distance map predictions via $\rm{L}_1$ loss, and 5) sky environment map predictions via $\rm{L}_1$ loss.
Then direct supervision $\mathcal{L}_{\rm{sup}2}$ for $\text{P-Net}$ is provided to local lighting predictions via $\rm{L}_1$ loss.
\begin{align}
  \vspace{-1mm}
  \mathcal{L}_{\rm{sup}1}~& = \|\tilde{\mathbf{A}}-\mathbf{A}\|_2+\|1-\tilde{\mathbf{N}}\cdot\mathbf{N}\|_2 +\|\tilde{\mathbf{G}}-\mathbf{G}\|_1 \nonumber\\
                        ~&+ \|\tilde{\mathbf{S}}-\mathbf{S}\|_2+\|\tilde{\mathbf{H}}_{\rm{g}}-\mathbf{H}_{\rm{g}}\|_1, \\
  \mathcal{L}_{\rm{sup}2} ~&=  \|\tilde{\mathbf{H}}_{\rm{l}}-\mathbf{H}_{\rm{l}}\|_1.
  \label{equ:loss_direct}
\end{align}
where the $\tilde{*}$ means the estimations of I-Net and $\cdot$ is the dot product for each vector in a matrix.

\vspace{2mm}
\noindent\textbf{Diffuse convolution loss.}
In order to guide the sky environment map estimated by I-Net to extract low-frequency lighting information from the encoded SH coefficients, 
we add a diffuse convolution loss $\mathcal{L}_{\rm{dif}}$ to force $\bH_{\rm{g}}$ applied with the diffuse convolution to have a close appearance with a relighted pure Lambertian surface from $\bP_{\rm{SH}}$:
\vspace{-1mm}
\begin{equation}
  \vspace{-1mm}
  \mathcal{L}_{\rm{dif}}=\frac{1}{N}\sum_{i=1}^{N}[D(\bH_{\rm{g}},i) - \text{diag}(\bold{\alpha_{o}})\mathbf{L}\bold{b}(\mathbf{N}_{\bold{ll}}(i))]^2,
  \label{equ:loss_diff}
\end{equation}
where $\bold{\alpha_o}=[1,1,1]^\top$ is the global diffuse albedo, $\mathbf{L} \in \mathbb{R}^{3\times9}$ is the SH coefficients by reshaping $\bP_{\rm{SH}}\in\mathbb{R}^{1\times27}$, $\mathbf{N}_{\bold{ll}}$ is the normal map of a sphere in panorama coordination and the second order SH basis is given by: $\bold{b}(\bold{n})=[1,n_x,n_y,n_z,3n_{z}^{2}-1,n_xn_y,n_xn_z,n_yn_z,n_x^{2}-n_y^{2}]^\top$. $D$ is the diffuse convolution function defined as
\vspace{-1mm}
\begin{equation}
  \vspace{-1mm}
D(\mathbf{H},i)=\frac{1}{K_i}\sum_{\omega\in\Omega_i}\bH(\omega)s(\omega)(\omega\cdot \bold{n}),
\label{equ:diff_conv}
\end{equation} 
where $\Omega_i$ is the hemisphere centered at pixel $i$ on the global lighting environment map, $\bold{n}$ is the normal vector at pixel $i$, and $K_i$ is the sum of solid angles on $\Omega_i$. $\omega$ is a unit vector of direction and $s(\omega)$ is the solid angle for a pixel in the panorama map of direction $\omega$ with different scale factors (because pixels in the panorama map at different latitudes correspond to projections on the unit sphere with different area sizes). 

\vspace{2mm}
\noindent\textbf{Inverse rendering reconstruction loss.}
To make the network learn constraints from physically-based image formation model, we put SH coefficients as an intermediate variable and provide indirect supervision to $\bP_{\rm{SH}}$ via an inverse rendering reconstruction loss $\mathcal{L}_{\rm{rec}}$ on the directly illuminated part, by multiplying a non-shadowed mask to disregard the effect of shadows: 
\begin{equation}
  \mathcal{L}_{\rm{rec}} = ||\mathbf{M_{ncs}}\odot(\mathbf{I}_{\rm{im}}-{(\tilde{\mathbf{A}}\odot\mathbf{L}\mathbf{B}(\tilde{\mathbf{N}})})^{1/\gamma})||_2,
  \label{equ:loss_rec}
\end{equation}
where $\odot$ represents the element-wise product. We use a fixed $\gamma=2.2$ to compress the dynamic range. $\mathbf{M_{ncs}}$ is the non-shadowed mask computed using shadow maps from intrinsics; a binary Otsu segmentation on histogram of shadow maps is further used to eliminate weak interreflections; $\mathbf{I}_{\rm{im}}\in \mathbb{R}^{3\times K}$ is the RGB image matrix; $\tilde{\mathbf{A}}\in \mathbb{R}^{3\times K}$ is the estimated diffuse albedo matrix; $\mathbf{B}(\tilde{\mathbf{N}})\in \mathbb{R}^{9\times K}$ is a matrix stacked by $\mathbf{b}(\mathbf{n})$ which applied SH basis on the normal map.

\vspace{2mm}
\noindent\textbf{Tonemapped SSIM loss.}
A structural similarity index measure ($\rm{SSIM}$) loss between dynamic range compressed images with a fixed gamma parameter ($2.2$ in our experiment) is used to recover structure similarity between the estimation and ground truth:
\begin{equation}
\mathcal{L}_{\rm{tom}}=\|\Im((2^{e}\cdot\tilde{\mathbf{H}}_{\rm{l}})^{1/\gamma})-\Im((2^{e}\cdot{\mathbf{H}}_{\rm{l}})^{1/\gamma})\|_{\rm{SSIM}}, \nonumber
\end{equation}
\begin{equation}
\quad\Im(\mathbf{H}) = \left\{ \begin{aligned} 
1 & & \mathbf{H} > 1 \\ 
0 & & \mathbf{H} < 0 \\
\mathbf{H} & &  0 \leq \mathbf{H} \leq1
\end{aligned} \right .
\label{equ:loss_tom}
\end{equation}
where $e$ is the exposure intensity fixed as $-0.3$ in our experiments.

\begin{figure}
  \centering
  \includegraphics[width=1\columnwidth]{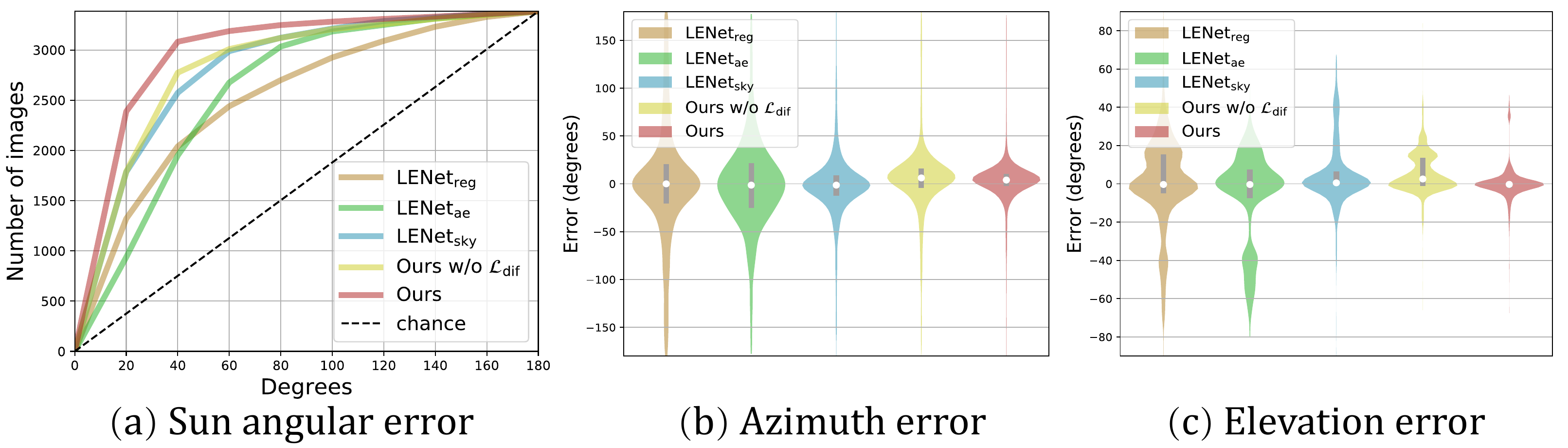}
  \caption{Quantitative evaluation of sun position estimation. (a) The cumulative sun angular error comparison between baseline methods and ours. The estimation error of sun azimuth (b) and elevation angles (c) is displayed as a ``violin plot" where the envelope of each bin represents the percentile, the gray line represents the percentile of 25$\%$ to 75$\%$, and the median is shown as a white point.}
  \label{fig:quan_sun_error}
  \vspace{-1.5mm}
\end{figure}

I-Net is trained by summing up direct supervision loss, diffuse convolution loss, and inverse rendering reconstruction loss as: $\mathcal{L}_{\rm{I}} = \mathcal{L}_{\rm{sup}1} + \mathcal{L}_{\rm{diff}} + \mathcal{L}_{\rm{rec}}$, and then P-Net is trained by summing up direct supervision loss and tonemapped $\text{SSIM}$ loss as: $\mathcal{L}_{\rm{P}} = \mathcal{L}_{\rm{sup}2}+ \mathcal{L}_{\rm{tom}}$.
\begin{figure}
  \centering
    \includegraphics[width=1\columnwidth]{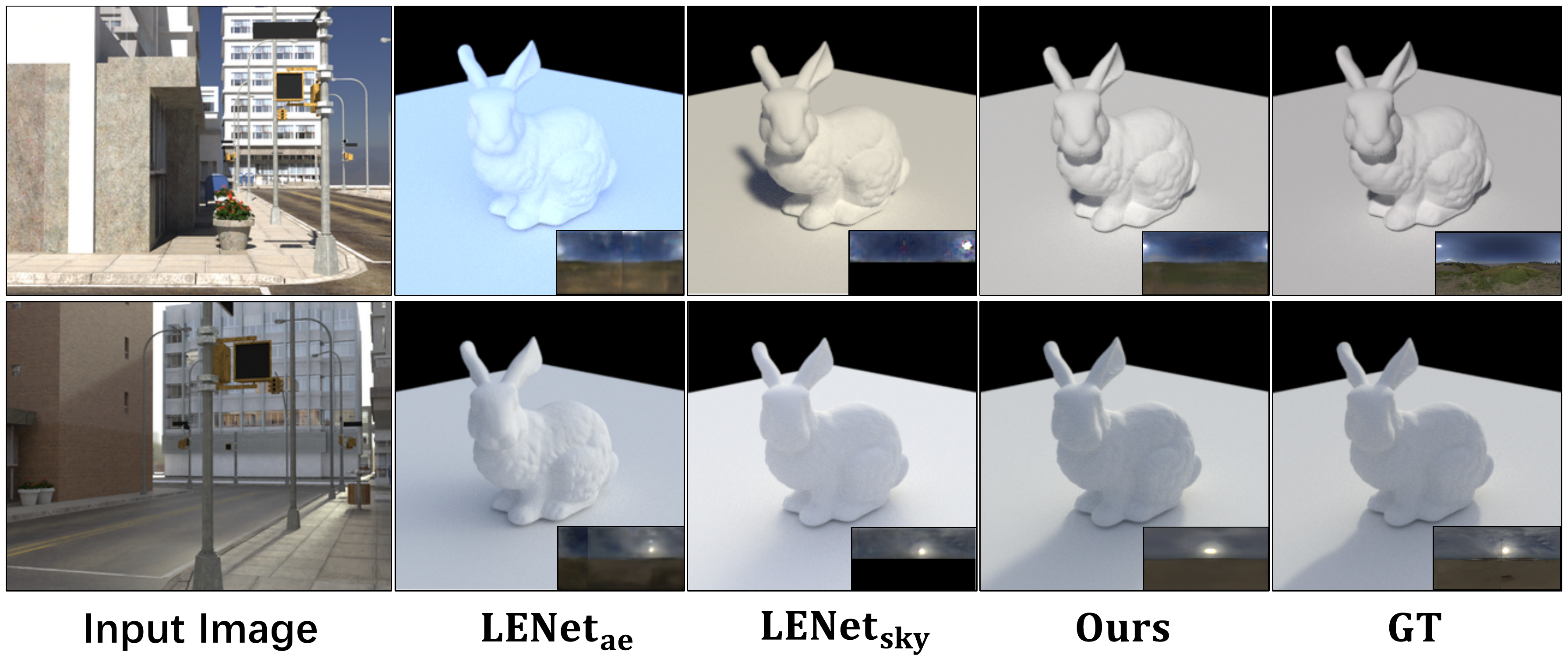}
    \caption{Relighting results with global lighting (shown as insets) on our SOLID-Img dataset.}
    \label{fig:relighting_results}
    \vspace{-2mm}
\end{figure}


\section{Experiments}

We perform detailed network analysis and present qualitative and quantitative results on our SOLID-Img test set. We also capture a small set of real LDR outdoor local environment maps to analyse the generalization of our method. Finally, we show relighted bunny results to validate our methods qualitatively\footnote{More results are in the supplementary material.}. 
To measure the accuracy of our predicted global sky environment map $\bH_{\rm{g}}$ and local illumination maps $\bH_{\rm{l}}$, we use mean absolute error (MAE) on the HDR sky environment map, angular error on the sun position and sun azimuth/elevation angles, and SSIM on the detailed local lighting as error metrics. 

\subsection{Analysis using Synthetic Dataset}
\vspace{1mm}
\noindent\textbf{Effectiveness of I-Net.}
To validate the design of intrinsic decomposition, we compare our global lighting estimation branch with three baseline models for the accuracy of estimated sun positions: 1) $\bold{LENet_{reg}}$ is a regression-based model that directly regresses the global sky from the input image. 2) $\bold{LENet_{ae}}$ is a two-stream convolution network used to regress sun azimuth angle and normalized HDR panorama from an LDR panorama~\cite{ldr2hdr}; we modify the input as a single limited-FOV image to adapt our task. 3) $\bold{LENet_{sky}}$ learns to estimate both the sun azimuth angle and a non-parametric sky~\cite{deepsky}. In particular, $\bold{LENet_{ae}}$ learns azimuth estimation as a regression task, while $\bold{LENet_{sky}}$ treats it as a classification problem. All baseline models are retrained using SOLID-Img training dataset with the same setting\footnote{Detailed model structures are in the supplementary material.}. Since our global lighting is represented by a non-parametric sky environment map, we compute the sun position by finding the largest connected component of the sky above a threshold (98\%) and computing its centroid. And then we rotate estimated sky environment maps around their azimuth angles to make sure the sun is in the center of the image so that we can compare it with their baseline models. 

From \Fref{fig:quan_sun_error}, we can see that our method shows significant improvement than $\bold{LENet_{reg}}$ and $\bold{LENet_{ae}}$ and comparable improvement than $\bold{LENet_{sky}}$, thanks to the intrinsic cues. Qualitative results on the test dataset are shown in~\Fref{fig:relighting_results}\footnote{Numerical results and MAE errors on estimated sky environment map are provided in the supplementary material.}. Our relighting results and estimated lightings show a closer appearance to the ground truth (shown as insets) than other methods.

\begin{figure}
  \centering
    \includegraphics[width=1\columnwidth]{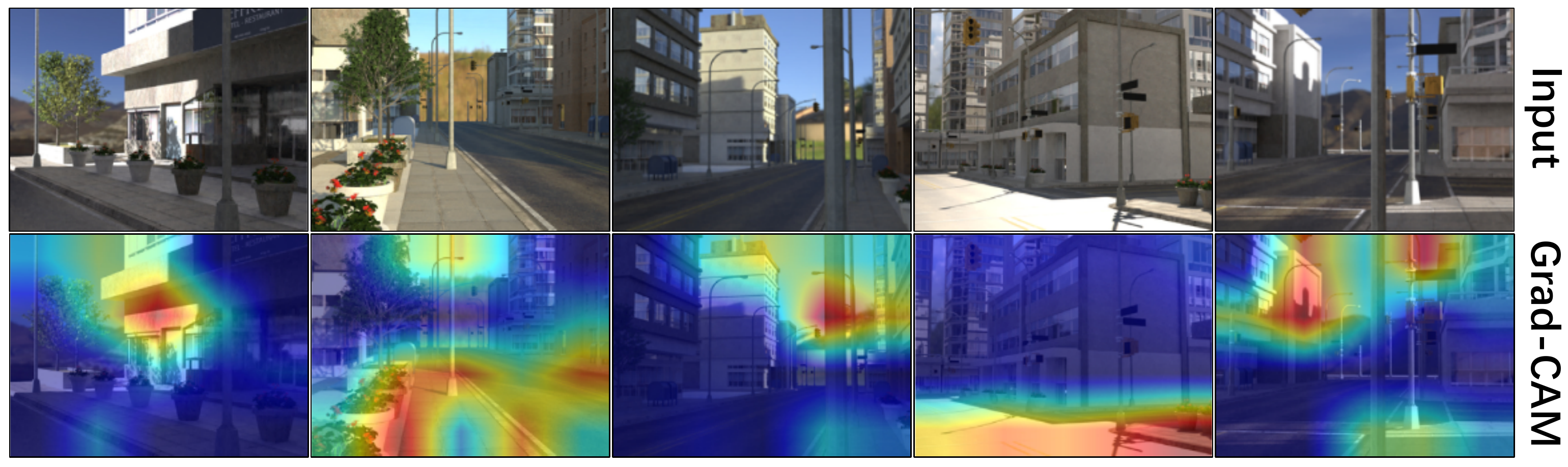}
    \caption{Visualization of Grad-CAM~\cite{gradcam} on our SH lighting prediction using SOLID-Img test set.}
    \label{fig:SH_features}
    \vspace{-1mm}
\end{figure}

To help understanding how SH coefficients decode the global lighting information, we perform the Grad-CAM~\cite{gradcam} on our global lighting encoder. We use the maximum response value of SH coefficients as the target backward label to find which regions of input are important for global lighting prediction. From~\Fref{fig:SH_features}, the feature heatmaps validate that I-Net mostly captures directly illuminated information to estimate global lighting.

\begin{figure}
  \centering
    \includegraphics[width=1\columnwidth]{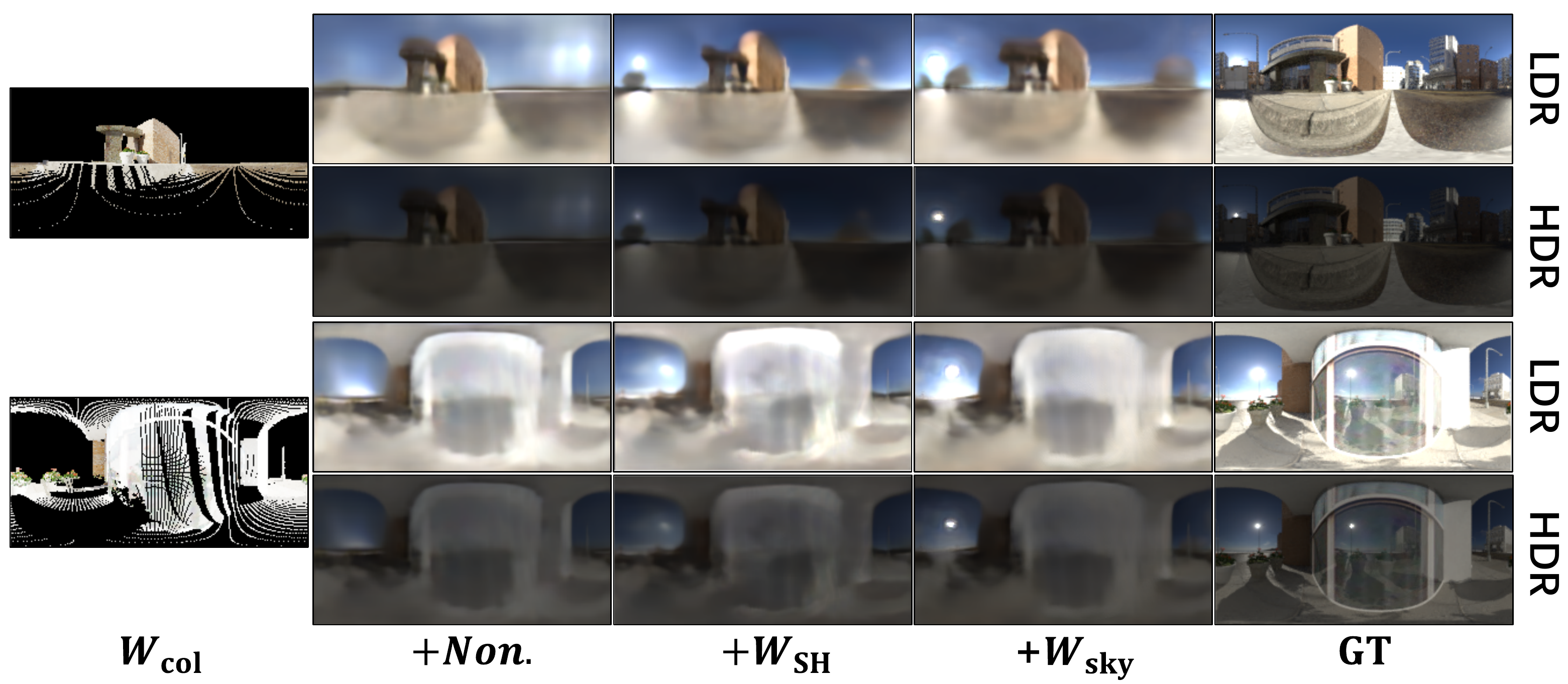}
    \caption{Estimated local lighting with different inputs.}
    \label{fig:qual_multi_input}
    \vspace{-2mm}
 \end{figure}

\begin{table}
  \centering
  \caption{Ablation study about our multi-input module.}  \label{tab:ablation_multi_input}
  \begin{threeparttable} 
     \setlength{\tabcolsep}{5.5mm}{
    \begin{tabular}{lcc}
    \toprule
    Inputs & SSIM & MAE \cr
    \midrule
    \text{$W_{\rm{col}}$} & $0.689$ & $0.848$ \\
    \text{$W_{\rm{col}}$ + $W_{\rm{sha}}$} & $0.736$ & $0.730$ \\
    \text{$W_{\rm{col}}$ + $W_{\rm{SH}}$} & $0.787$ & $0.618$ \\
    \text{$W_{\rm{col}}$ + $W_{\rm{sky}}$} & $0.793$ & $0.531$ \\
    \text{$W_{\rm{col}}$ +  $W_{\rm{sha}}$+ $W_{\rm{sky}}$} & $\textbf{0.803}$ & $\textbf{0.523}$ \\
    \bottomrule
    \end{tabular}}
  \end{threeparttable}
\end{table}
\vspace{2mm}
\noindent\textbf{Effectiveness of P-Net.}
We train our $\text{P-Net}$ with combinations of different inputs: warped incomplete LDR image panorama $W_{\rm{col}}$, relighted Lambertian surface $W_{\rm{SH}}$, estimated sky environment map
$W_{\rm{sky}}$, and incomplete shadow panorama image $W_{\rm{sha}}$. During training, we only process direct supervision on local lighting. We evaluate the SSIM and MAE errors between the estimated
local lighting and ground truth.  From~\Tref{tab:ablation_multi_input}, we can tell that directly providing $W_{\rm{sky}}$ rather than $W_{\rm{SH}}$ improves our algorithm marginally, while also
providing $W_{\rm{sha}}$ improves it a bit more. We conjecture it is because shadows provide occlusion information which is helpful for lighting estimation. In~\Fref{fig:qual_multi_input}, we show
results without global lighting, with $W_{\rm{SH}}$ as global lighting, and with $W_{\rm{sky}}$ as global lighting, respectively. We find that P-Net is incapable to learn the correct sun position only
with the warped color image but can recover it accurately by adding $W_{\rm{SH}}$ as shown in the first column and third column. Although the sun position is well recovered with $W_{\rm{SH}}$, the sun
intensity still has a large gap from the real condition.

\begin{figure}
  \centering
    \includegraphics[width=1\columnwidth]{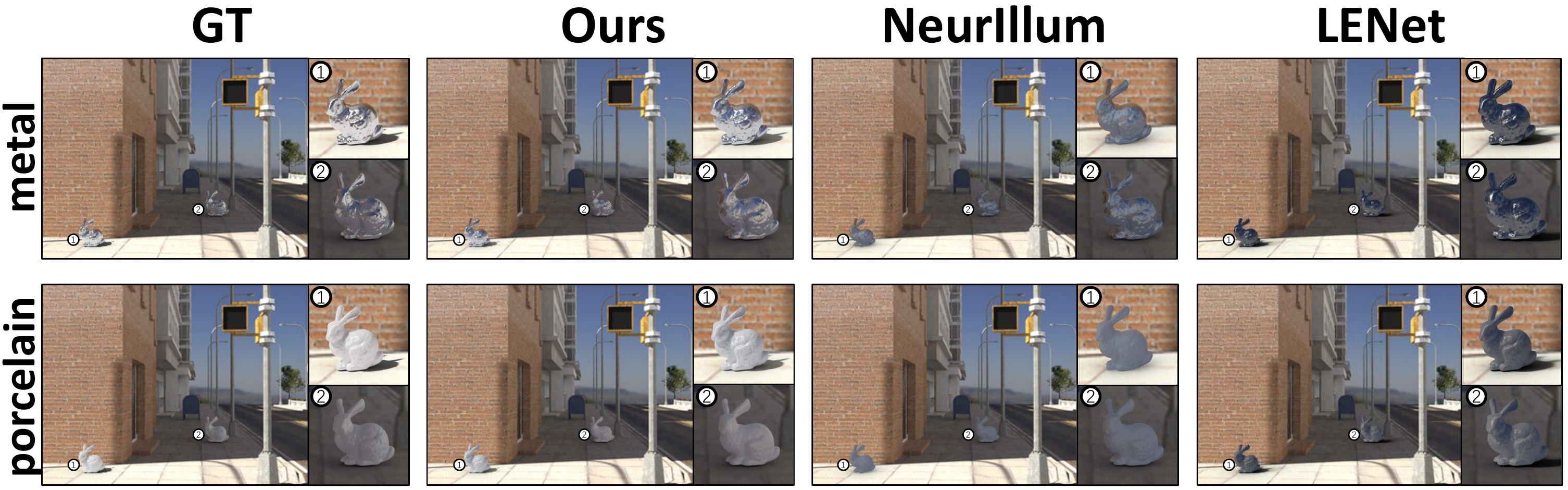}
    \caption{\edwardzhu{Synthetic examples of inserting virtual objects of different materials, compared with NeurIllum~\cite{indoorlocallighting_pano} and LENet~\cite{deepsky}.}} 
    \label{fig:supp-syn-insertion}
    \vspace{-2mm}
\end{figure}

\edwardzhu{For an off-the-shelf renderer (\eg, Blender), we can achieve multi-object rendering by setting it to render only the object in the selected lighting position, and then blending this result
  with rendering results from other positions through the alpha channel. In~\Fref{fig:supp-syn-insertion}, we show the visual quality of synthetic object insertion to better illustrate the usefulness of spatially-varying outdoor lighting estimation. As can be observed, our method can render correct lighting effects (specular highlights and shadows) on rabbits under different materials.} 

\vspace{2mm}
\noindent\textbf{Effects of different losses.}
To verify the necessity of each loss function, we evaluate the performance of $\text{I-Net}$ and $\text{P-Net}$ using different combinations of loss functions.  In~\Fref{fig:quan_sun_error}, we can
observe that our models have comparable improvement to $\bold{LENet_{sky}}$ even without $\mathcal{L}_{\rm{dif}}$ due to constraint from intrinsics provided by $\text{I-Net}$. By further adding
$\mathcal{L}_{\rm{dif}}$, $\text{I-Net}$ can learn the global sky environment map more effectively with the guidance of SH encoded lighting and then produce a more accurate sky estimation than
$\bold{LENet_{sky}}$. If $\mathcal{L}_{\rm{tom}}$ is ablated, the performance on test dataset becomes $0.760$ / $0.564$ ($\text{SSIM}$ / $\text{MAE}$); by adding this loss, the numbers are
$\textbf{0.798}$ / $\textbf{0.552}$ ($\text{SSIM}$ / $\text{MAE}$), which shows that $\text{P-Net}$ can predict local lighting more accurately especially on structure similarity.

\begin{figure*}
  \centering
    \includegraphics[width=2.1\columnwidth]{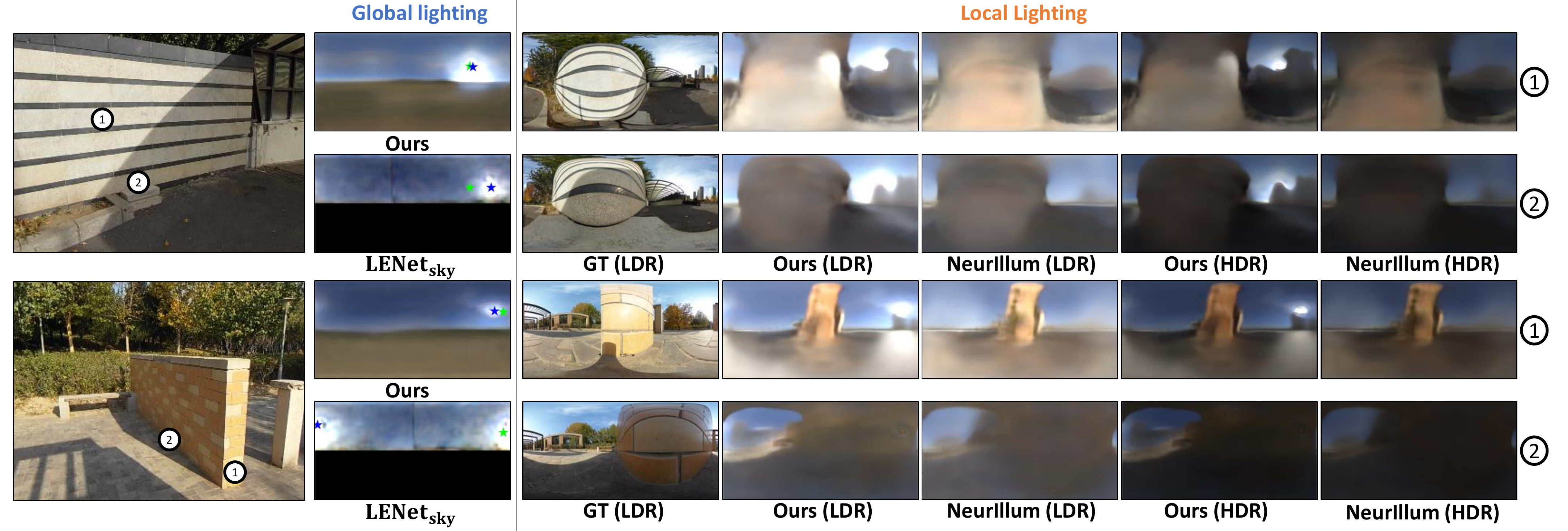}
    \caption{Comparison of estimated global and local lighting on our real test dataset. Column 1 shows the input image and selected pixel locations. Column 2 shows the estimated global lighting of DeepSky model~\cite{deepsky} and our method (blue star marks the estimated sun position by computing the centroid of largest connected component, while the green star marks the ground truth sun position labeled from a low-exposure environment map by us manually). Column 3 shows the LDR local lighting environment map. Columns 4$-$7 show the estimated local lighting by NeurIllum~\cite{indoorlocallighting_pano} and our methods in both LDR and HDR formats.}
    \label{fig:real_results}
    \vspace{-3mm}
\end{figure*}
\begin{figure}
  \centering
    \includegraphics[width=1\columnwidth]{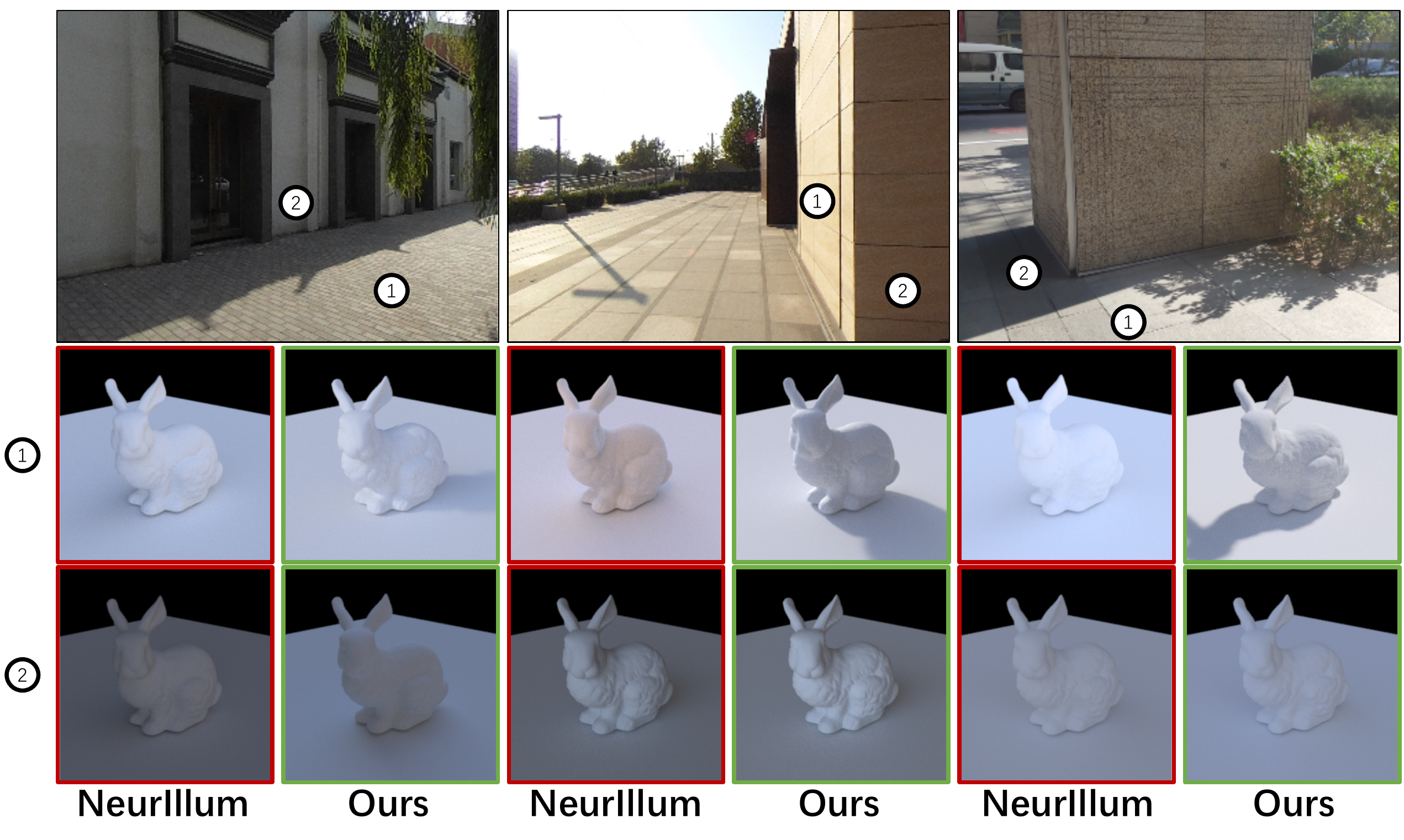}
    \caption{Qualitative comparison of relighting results using our real dataset.} 
    \label{fig:real_relight}
    \vspace{-5mm}
\end{figure}

\subsection{Evaluation on Real Dataset}
\vspace{2mm}
\noindent\textbf{Real data capture.}
To validate that SOLID-Net is able to perform outdoor local lighting estimation, we capture real outdoor city street view scenes and the corresponding spatially-varying local environment maps (see~\Fref{fig:real_results}). The images are captured by a Ricoh Theta SC2 camera with dual fisheye lens. For local lighting environment maps, the scenes are captured 1/2500s shutter speed with $f$2.0 aperture by placing the panoramic camera as a light probe at different locations. Due to the limited dynamic range of our panoramic camera, the local environment maps are not able to faithfully record the intensity of sunlight. To obtain the accurate sun position for evaluation, we further capture a low-exposure panorama with 1/25000s shutter speed and label the sun position manually. The captured LDR local lighting is aligned to its view vector with respect to the camera facing direction. In total, our real test dataset includes 29 outdoor scenes and 67 LDR local lighting environment maps for evaluating our method quantitatively.

\vspace{2mm}
\noindent\textbf{Comparison with previous work.}
We first compare the accuracy of global lighting estimation with $\bold{LENet_{sky}}$ model~\cite{deepsky} using sun position errors. The azimuth/elevation angular errors of $\bold{LENet_{sky}}$ are $37.5^\circ/9.5^\circ$. In contrast, our method maintains a high accuracy with $\textbf{28.9}^\circ/\textbf{6.8}^\circ$. From Column 2 of~\Fref{fig:real_results}, we can see that our method can generate a clearer environment map with different sky conditions and our estimated sun positions are closer to the ground truth than $\bold{LENet_{sky}}$.  To evaluate the estimated local lighting, we compare our method with NeurIllum~\cite{indoorlocallighting_pano} retrained on our synthetic dataset on estimated spatially-varying lighting quantitatively and qualitatively. Overall, our method achieves a better SSIM / MAE (the higher is better / the lower is better) performance of $\textbf{0.235}$ / $\textbf{0.203}$, compared to $0.228$ / $0.233$ for NeurIllum. Compared estimations of our method (Column 4-5) and NeurIllum (Column 6-7) with ground truth (Column 3) in~\Fref{fig:real_results},  we note that their method does not capture the accurate sun position and intensity, caused by missing panoramic information which our method handles well. We also show relighted bunny results to further compare estimated spatially-varying lighting effects of our method and NeurIllum (see~\Fref{fig:real_relight}). These show that our approach adapts to strongly spatially-varying local lighting effects in real scenes.
 


\vspace{-3mm}
\begin{figure}
  \centering
    \includegraphics[width=1\columnwidth]{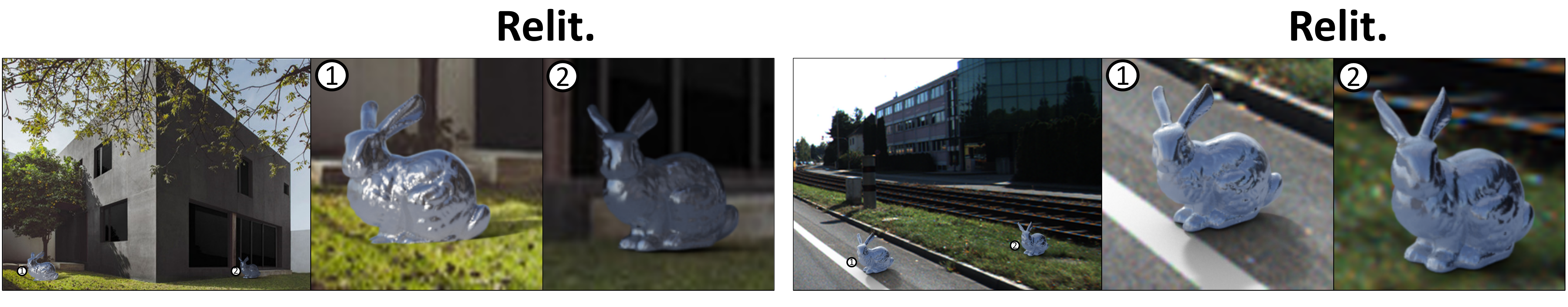}
    \caption{\edwardzhu{Real examples of virtual object insertion.}}
    \label{fig:stress-test}
    \vspace{-3mm}
\end{figure}

\begin{figure}
  \centering
    \includegraphics[width=1\columnwidth]{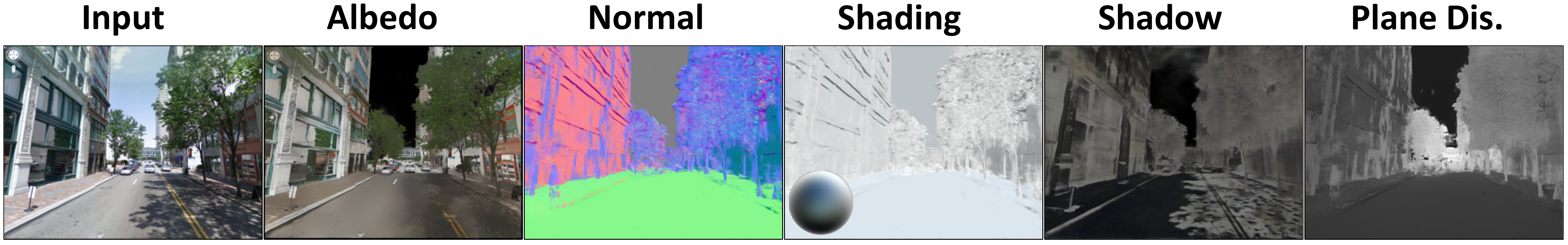}
    \caption{\edwardzhu{Real examples of intrinsic decomposition.}}
    \label{fig:supp-intrinsic-real}
    \vspace{-3mm}
\end{figure}

\section{Discussion}

\vspace{-1mm}
\edwardzhu{We present the first end-to-end outdoor spatially-varying lighting estimation framework and demonstrate it significantly outperforms previous works via extensive evaluations on both synthetic and real datasets. Our method is able to generalize on real scenes with a slightly different appearance from our synthetic scene. An example is shown in~\Fref{fig:stress-test}, in which the virtual object is reasonably relit in a scene of rarely seen structures (with railway and glass) in the synthetic training data.}

\vspace{2mm}
\noindent\textbf{Limitations and future work.}
\edwardzhu{Due to the material diversity gap between synthetic and real data, the intrinsic decomposition results on real data may not be as accurate as those on synthetic data
  (\Fref{fig:supp-intrinsic-real} compared with~\Fref{fig:synthetic_results}). Although SOLID-Net estimates HDR lighting environment map to support realistic relighting effects, our lighting model is
  not suitable for generating animations that are sensitive at harsh lighting boundaries, which will be an interesting direction for future work.}
{\small
\bibliographystyle{ieee_fullname}
\bibliography{OutdoorLighting}

\begin{thebibliography}{10}\itemsep=-1pt

\bibitem{blender}
{Blender.}
\newblock \url{https://www.blender.org}.

\bibitem{scenecity}
{Blender SceneCity.}
\newblock \url{https://www.cgchan.com/store/scenecity}.

\bibitem{hdrihaven}
{HDRI HAVEN.}
\newblock \url{https://hdrihaven.com}.

\bibitem{Matterport3D}
Chang Angel, Angela Dai, Thomas Funkhouser, Maciej Halber, Matthias Niessner,
  Manolis Savva, Shuran Song, Andy Zeng, and Yinda Zhang.
\newblock {Matterport3D: Learning from RGB-D Data in Indoor Environments}.
\newblock {\em International Conference on 3D Vision (3DV)}, 2017.

\bibitem{Barron2013}
Jonathan~T. Barron and Jitendra Malik.
\newblock {Intrinsic Scene Properties from a Single RGB-D Image}.
\newblock In {\em Proc. of Computer Vision and Pattern Recognition}, 2013.

\bibitem{cheng2018}
Dachuan Cheng, Xiaoming~Deng Jian~Shi, Yanyun~Chen, and Xiaopeng Zhang.
\newblock {Learning Scene Illumination by Pairwise Photos from Rear and Front
  Mobile Cameras}.
\newblock {\em Computer Graphics Forum}, 37:213--221, 2018.

\bibitem{paul98}
Paul Debevec.
\newblock {Rendering Synthetic Objects into Real Scenes: Bridging Traditional
  and Image-based Graphics with Global Illumination and High Dynamic Range
  Photography}.
\newblock In {\em Proc. of ACM SIGGRAPH}, 1998.

\bibitem{lalonde17}
Marc-Andr{\'e} Gardner, Kalyan Sunkavalli, Ersin Yumer, Xiaohui Shen, Emiliano
  Gambaretto, Christian Gagn{\'e}, and Jean-Fran{\c{c}}ois Lalonde.
\newblock {Learning to Predict Indoor Illumination from a Single Image}.
\newblock In {\em Proc. of ACM SIGGRAPH Asia}, 2017.

\bibitem{indoorlocallighting_SH}
Mathieu Garon, Kalyan Sunkavalli, Sunil Hadap, Nathan Carr, and
  Jean-Fran\c{c}ois Lalonde.
\newblock {Fast Spatially-Varying Indoor Lighting Estimation}.
\newblock In {\em Proc. of Computer Vision and Pattern Recognition}, 2019.

\bibitem{deepsky}
Yannick Hold-Geoffroy, Akshaya Athawale, and Jean-Fran{\c{c}}ois Lalonde.
\newblock {Deep Sky Modeling for Single Image Outdoor Lighting Estimation}.
\newblock In {\em Proc. of Computer Vision and Pattern Recognition}, 2019.

\bibitem{outdoorlighting_param}
Yannick Hold-Geoffroy, Emiliano~Gambaretto Kalyan~Sunkavalli, Sunil~Hadap, and
  Jean-Fran\c{c}ois Lalonde.
\newblock {Deep Outdoor Illumination Estimation}.
\newblock In {\em Proc. of Computer Vision and Pattern Recognition}, 2017.

\bibitem{lalonde12}
Alexei A.~Efros Jean-Fran\c{c}ois~Lalonde and Srinivasa~G. Narasimhan.
\newblock {Estimating the Natural Illumination Conditions from a Single Outdoor
  Image}.
\newblock {\em International Journal of Computer Vision}, 98(2):123--145, 2012.

\bibitem{adam}
Diederik~P Kingma and Jimmy Ba.
\newblock Adam: A method for stochastic optimization.
\newblock 2015.

\bibitem{skydatabase}
JF Lalonde, LP Asselin, J Becirovski, Y Hold-Geoffroy, M Garon, MA Gardner, and
  J Zhang.
\newblock {The Laval HDR sky database}, 2016.
\newblock \url{http://sky.hdrdb.com}.

\bibitem{legendre19}
Chloe LeGendre, Graham~Fyffe Wan-Chun~Ma, John Flynn~Laurent Charbonnel, Jay
  Busch, and Paul Debevec.
\newblock {DeepLight: Learning Illumination for Unconstrained Mobile Mixed
  Reality}.
\newblock In {\em Proc. of Computer Vision and Pattern Recognition}, 2019.

\bibitem{indoorlighting_inverserender}
Zhengqin Li, Mohammad Shafiei, Ravi Ramamoorthi, Kalyan Sunkavalli, and
  Manmohan Chandraker.
\newblock {Inverse rendering for complex indoor scenes: Shape,
  spatially-varying lighting and svbrdf from a single image}.
\newblock In {\em Proc. of Computer Vision and Pattern Recognition}, 2020.

\bibitem{pytorch}
Adam Paszke, Sam Gross, Francisco Massa, Adam Lerer, James Bradbury, Gregory
  Chanan, Trevor Killeen, Zeming Lin, Natalia Gimelshein, Luca Antiga, et~al.
\newblock Pytorch: An imperative style, high-performance deep learning library.
\newblock In {\em Proc. of Neural Information Processing Systems}, 2019.

\bibitem{unet}
Olaf Ronneberger, Philipp Fischer, and Thomas Brox.
\newblock {U-Net: Convolutional Networks for Biomedical Image Segmentation}.
\newblock In {\em Medical Image Computing and Computer-Assisted Intervention},
  2015.

\bibitem{gradcam}
Ramprasaath~R. Selvaraju, Abhishek~Das Michael~Cogswell, Devi~Parikh
  Ramakrishna~Vedantam, and Dhruv Batra.
\newblock {Grad-CAM: Visual Explanations from Deep Networks via Gradient-based
  Localization}.
\newblock In {\em Proc. of International Conference on Computer Vision}, 2017.

\bibitem{indoorlocallighting_pano}
Shuran Song and Thomas Funkhouser.
\newblock {Neural Illumination: Lighting Prediction for Indoor Environments}.
\newblock In {\em Proc. of Computer Vision and Pattern Recognition}, 2019.

\bibitem{suncg}
Shuran Song, Fisher Yu, Andy Zeng, Angel~X Chang, Manolis Savva, and Thomas
  Funkhouser.
\newblock {Semantic Scene Completion from a Single Depth Image}.
\newblock In {\em Proc. of Computer Vision and Pattern Recognition}, 2017.

\bibitem{im2pano3d}
Shuran Song, Andy Zeng, Angel~X Chang, Manolis Savva, Silvio Savarese, and
  Thomas Funkhouser.
\newblock {Im2Pano3D: Extrapolating 360 Structure and Semantics Beyond the
  Field of View}.
\newblock In {\em Proc. of Computer Vision and Pattern Recognition}, 2019.

\bibitem{paul04}
Jessi Stumpfel, Andreas~Wenger Andrew~Jones, Tim~Hawkins Chris~Tchou, and Paul
  Debevec.
\newblock {Direct HDR Capture of the Sun and Sky}.
\newblock In {\em Proc. of ACM SIGGRAPH}, 2004.

\bibitem{inverserendernet}
Ye Yu and William~AP Smith.
\newblock {InverseRenderNet: Learning single image inverse rendering}.
\newblock In {\em Proc. of Computer Vision and Pattern Recognition}, 2019.

\bibitem{googlestreetview}
Amir~Roshan Zamir and Mubarak Shah.
\newblock {Image Geo-localization Based on Multiple Nearest Neighbor Feature
  Matching using Generalized Graphs}.
\newblock {\em IEEE Transactions on Pattern Analysis and Machine Intelligence},
  PP(99):1--1, 2014.

\bibitem{allweather}
Jinsong Zhang, Yannick Hold-Geoffroy Kalyan~Sunkavalli, Jonathan~Eisenmann
  Sunil~Hadap, and Jean-François Lalonde.
\newblock {All-Weather Deep Outdoor Lighting Estimation}.
\newblock In {\em Proc. of Computer Vision and Pattern Recognition}, 2019.

\bibitem{ldr2hdr}
Jinsong Zhang and Jean-Fran\c{c}ois Lalonde.
\newblock {Learning High Dynamic Range from Outdoor Panoramas}.
\newblock In {\em Proc. of International Conference on Computer Vision}, 2017.

\bibitem{physicallyrender}
Yinda Zhang, Shuran Song, Ersin Yumer, Manolis Savva, Joon-Young Lee, Hailin
  Jin, and Thomas Funkhouser.
\newblock {Physically-Based Rendering for Indoor Scene Understanding Using
  Convolutional Neural Networks}.
\newblock In {\em Proc. of Computer Vision and Pattern Recognition}, 2017.

\end{thebibliography}
}


\newpage
\setcounter{section}{0}
\setcounter{figure}{0}


\maketitle

In this supplementary material, we provide more details about our implementation details, synthetic data generation, network architectures, and additional results on synthetic data and real images in the wild.

\begin{figure*}
  \centering
    \includegraphics[width=2\columnwidth]{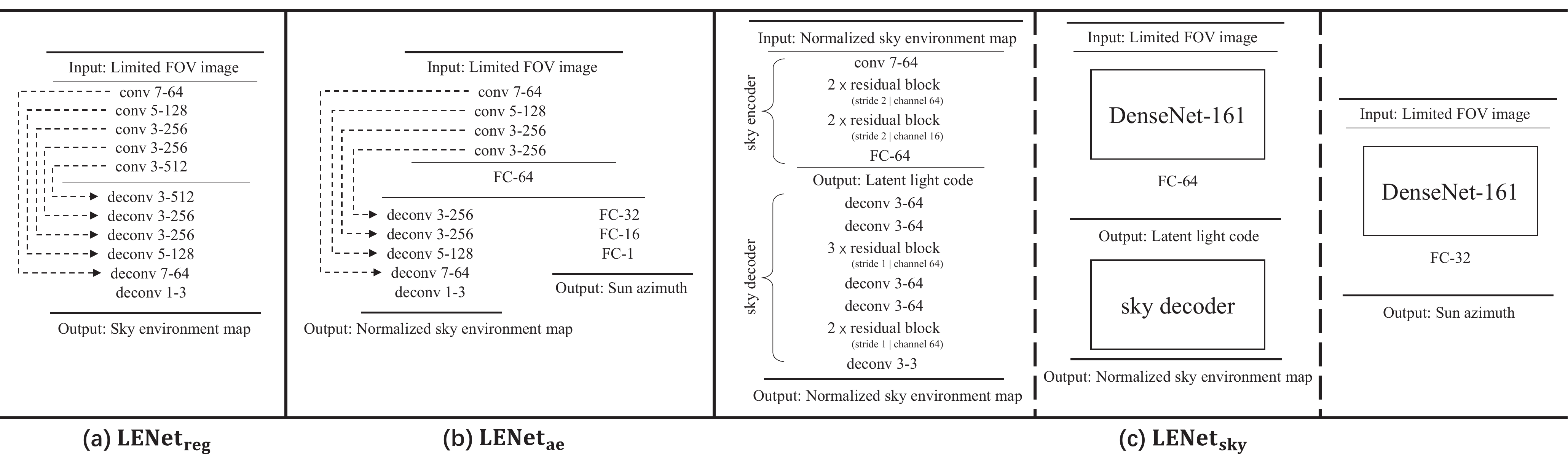}
    \caption{Structures of baseline models.} 
    \label{fig:baseline}
\end{figure*}

\section{Implementation Details}
To train SOLID-Net, we use SOLID-Img augmented with random flip and crop. Our framework is implemented in PyTorch~\cite{pytorch} and Adam~\cite{adam} optimizer is used with default parameters. We first train I-Net using a batch size of 8 for 20 epochs until convergence, and then train P-Net with a batch size of 4 for 60 epochs on an RTX2080 GPU. We find that an end-to-end fine-tuning does not improve the performance. The learning rate is initially set to $5\times 10^{-4}$ and halved every 5 epochs for both networks. Training convergence takes roughly 24 hours.

\section{Additional Details in Data Generation}
The data generation pipeline for \textbf{SOLID-Img} dataset has been introduced in Section 3.1 of the main paper. Here, we introduce additional details about the \textbf{camera setting} step. 

For each road block, we select a set of cameras with diverse views seeing most objects in the context, to provide comprehensive information for lighting estimation, as shown in Figure 2(a) of the main paper. Our process starts by selecting the ``best'' camera~\cite{physicallyrender} for each of the six horizontal view direction sectors in every road block. For each horizontal view direction, we sample a dense set of cameras on a 2D grid with $1$m solution, choosing a random camera viewpoint within each grid cell, a random horizontal view direction within the $60^\circ$ sector, a random height of $1.55\pm 0.05$m above the floor uniformly, and a pitch angle within $10^\circ$ around horizontal direction. Then for each camera, we render an item buffer and count the number of visible pixels according to Z-buffer and the number of objects. For each horizontal view direction in each road block, we select the view with the highest percentage pixel coverage, as long as it has more than three object categories. 

\section{Details about Network Architectures}
In this section, we introduce the detailed network architectures of baseline models as shown in~\Fref{fig:baseline}.

The first baseline model, denoted as $\bold{LENet_{reg}}$, is a regression model that directly regresses the global sky environment map from the input limited-FOV image. The second baseline model, denoted as $\bold{LENet_{ae}}$, is a two-stream convolution network used to estimate sun position and normalized HDR panorama from a LDR panorama~\cite{ldr2hdr}; we modify the input as a single limited-FOV LDR image to adapt our task. The last baseline model, denoted as $\bold{LENet_{sky}}$, learns to estimate both the sun azimuth and the sky parameters from two image encoders and uses an autoencoder to learn the space of outdoor lighting by compressing an HDR sky image to a 64-dimensional latent vector and reconstructing it to a HDR sky environment map~\cite{deepsky}. In particular, $\bold{LENet_{ae}}$ learns azimuth estimation as a regression task, while $\bold{LENet_{sky}}$ treats it as a classification task. 

\begin{table}
  \centering
   \caption{Numerical results and MAE errors on estimated sky environment map.} \label{tab:ablation_inverse_rendering}
  \begin{threeparttable}  
     \setlength{\tabcolsep}{3mm}{
    \begin{tabular}{lccc}
    \toprule
    Methods & \text{$\xi_{azimuth}$} & \text{$\xi_{elevation}$} & $\text{$\xi_{\rm{HDR}}$}$ \cr
    \midrule
    $\bold{LENet_{reg}}$ & $37.0^\circ$ & $16.2^\circ$ & $0.508$  \\
    $\bold{LENet_{ae}}$~\cite{ldr2hdr}  & $34.0^{\circ}$ & $16.0^{\circ}$ & $0.542$  \\
    $\bold{LENet_{sky}}$~\cite{deepsky} &  $22.3^{\circ}$ & ${11.0}^{\circ}$ & $0.609$ \\
    $\textbf{Ours}$~($w/o$~$\mathcal{L}_{\rm{dif}}$) & ${19.1}^{\circ}$ & $11.4^{\circ}$ & ${0.491}$ \\
    $\textbf{Ours}$ & $\textbf{12.6}^{\circ}$ & $\textbf{8.5}^{\circ}$ & $\textbf{0.478}$ \\
    \bottomrule
    \end{tabular} }
  \end{threeparttable}
\end{table}

Numerical results about these baseline models compared with ours are shown in~\Tref{tab:ablation_inverse_rendering} ($\text{$\xi_{azimuth}$}$ is the sun azimuth angular error, $\text{$\xi_{elevation}$}$ is the sun elevation angular error, and $\text{$\xi_{\rm{HDR}}$}$ is the MAE error of normalized sky environment map). 
Training a single model to estimate a sky environment map with azimuth rotation is proved to be difficult (see Row 1 in~\Tref{tab:ablation_inverse_rendering}). By separating this task into the estimation of a normalized sky environment map (sun in the middle position along the horizontal direction) and sun azimuth angle, the results of single model lighting estimation get improved (see Row 2 in~\Tref{tab:ablation_inverse_rendering}). Further using three sub models to solve the whole problem results in more accurate estimation than using single models (see Row 3 in~\Tref{tab:ablation_inverse_rendering}). But the sun position and sky environment map are closely related, we prove that the estimation accuracy to both could be improved by our jointly training with intrinsic constraints being integrated in deep models (see Row 4-5 in~\Tref{tab:ablation_inverse_rendering}). 

\section{Qualitative Results on Synthetic Data}
More results of image decomposition and lighting estimation using SOLID-Img test dataset are shown in~\Fref{fig:synthetic_results}, \Fref{fig:relighting_results_global}, and~\Fref{fig:relighting_results_local}. Given an input image, our estimated albedo, normal, plane distance, shadow, and shading show close appearance to the ground truth (shown as insets) as shown in~\Fref{fig:synthetic_results}. In~\Fref{fig:relighting_results_global}, we can see that global lighting estimation resuls of SOILD-Net are closer to ground truth than three baseline models (we rotate the lighting of $\bold{LENet_{ae}}$ and $\bold{LENet_{sky}}$ according to the sun azimuth for better comparison). In addition, the relighted bunnies using our estimated lighting display accurate cast shadows, while other models fail to render.

As shown in~\Fref{fig:relighting_results_local}, our method can recover more accurate local lighting than NeurIllum~\cite{indoorlocallighting_pano} even for the reflection of the ground and some unseen parts (typical examples could be found in local lighting 1 in row 1, local lighting 3 in row 4, and local lighting 4 in row 5). We conjecture this is because our multi-input (global lighting for lighting information, shadow for occlusion information) module provides more useful information.

\section{Qualitative Results on Real Data}
More results of local lighting estimation on real dataset are shown in~\Fref{fig:relighting_results_local_real}. Compared with NeurIllum~\cite{indoorlocallighting_pano}, the lighting estimation results of our method are more similar to the ground truth in terms of overall structure, and our relighted bunnies show more realistic rendering apperances. 

\begin{figure*}
   \centering
     \includegraphics[width=2\columnwidth]{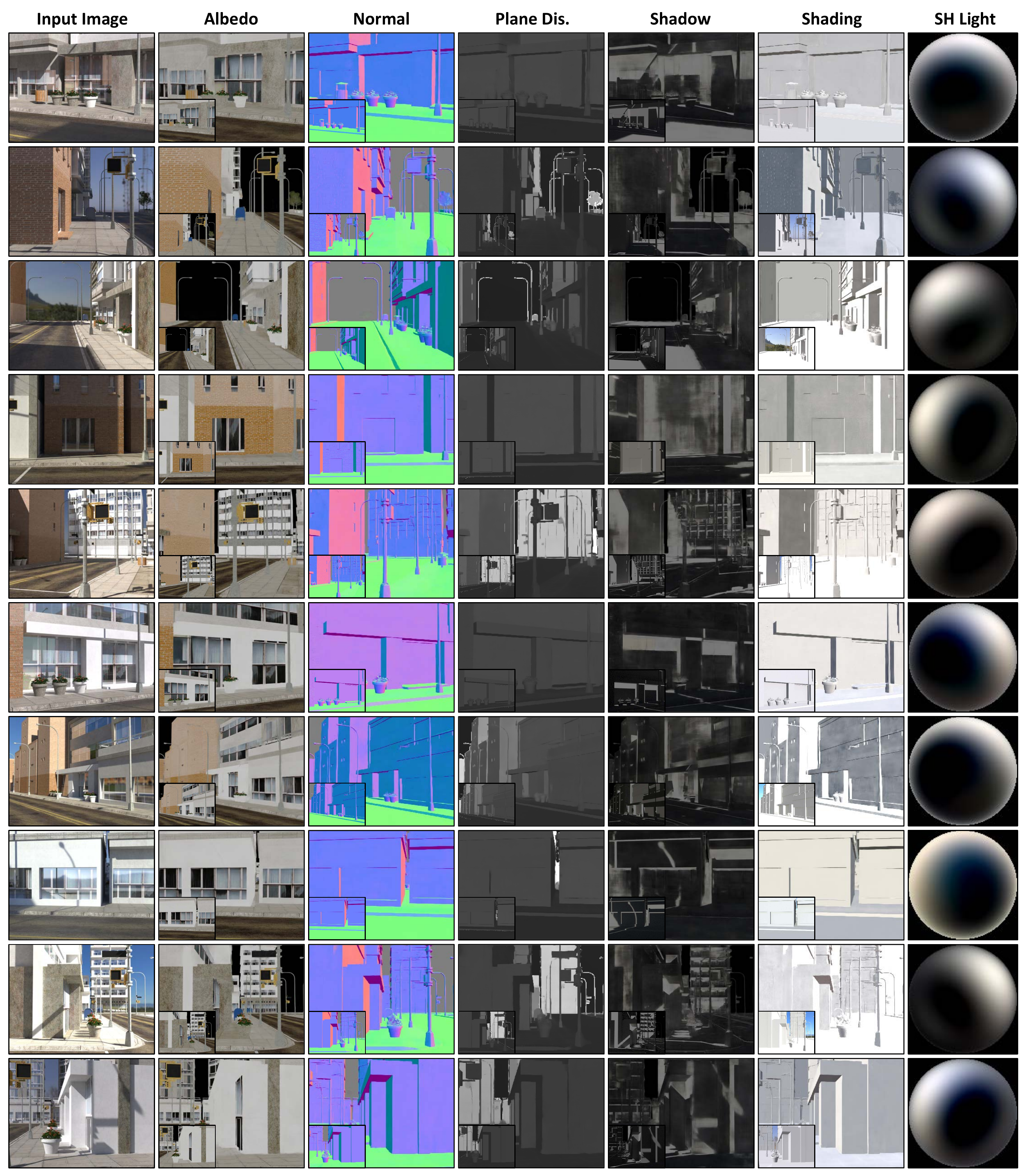}
     \caption{Intrinsic decomposition results and ground truth (shown as insets) on SOLID-Img test dataset.} 
     \label{fig:synthetic_results}
 \end{figure*}

 \begin{figure*}
   \centering
     \includegraphics[width=2\columnwidth]{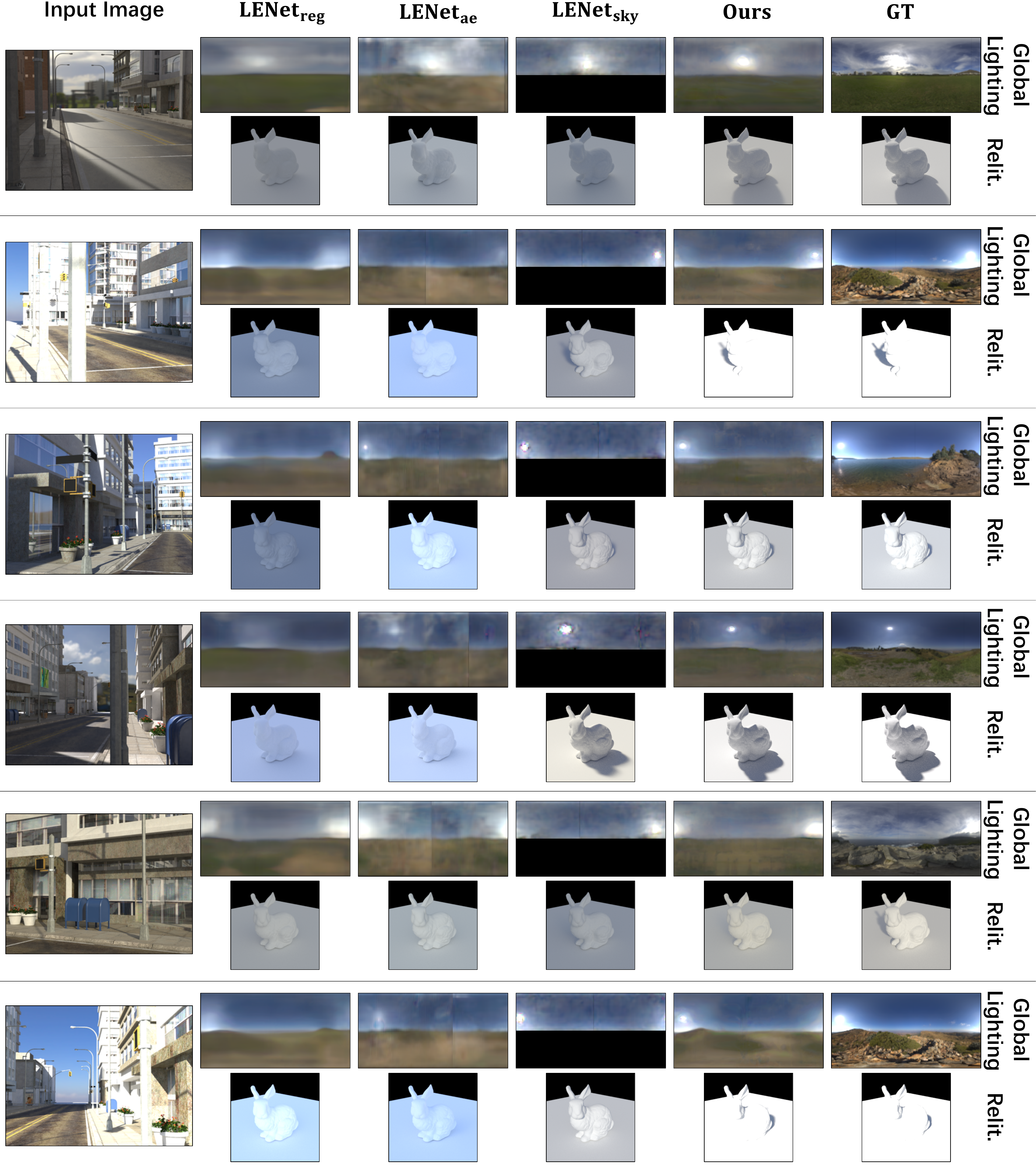}
     \caption{Global lighting estimation and relighting results on SOLID-Img test dataset.}
     \label{fig:relighting_results_global}
 \end{figure*}

 \begin{figure*}
   \centering
     \includegraphics[width=2\columnwidth]{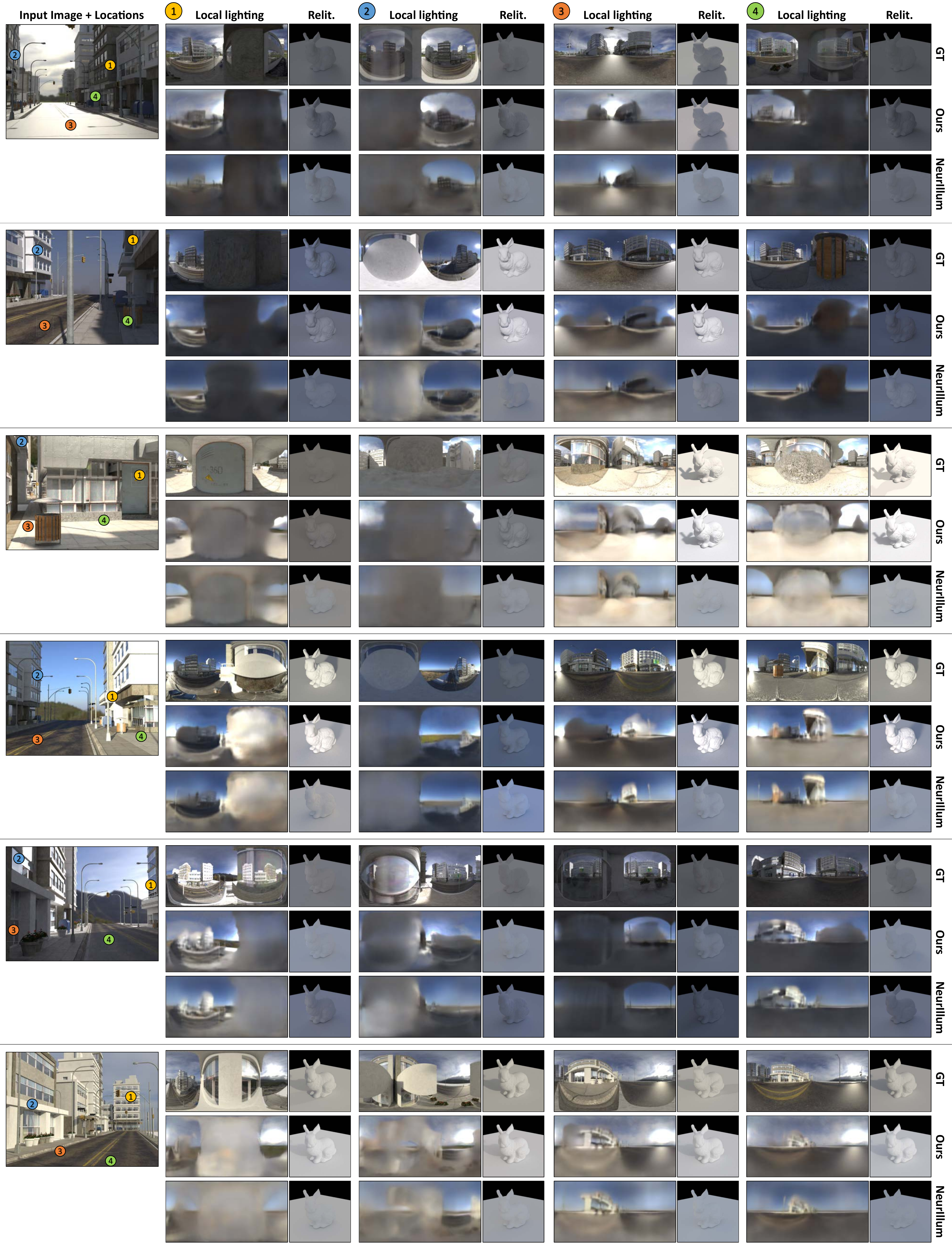}
     \caption{Local lighting estimation and relighting results on SOLID-Img test dataset.}
     \label{fig:relighting_results_local}
 \end{figure*}

 \begin{figure*}
   \centering
     \includegraphics[width=2\columnwidth]{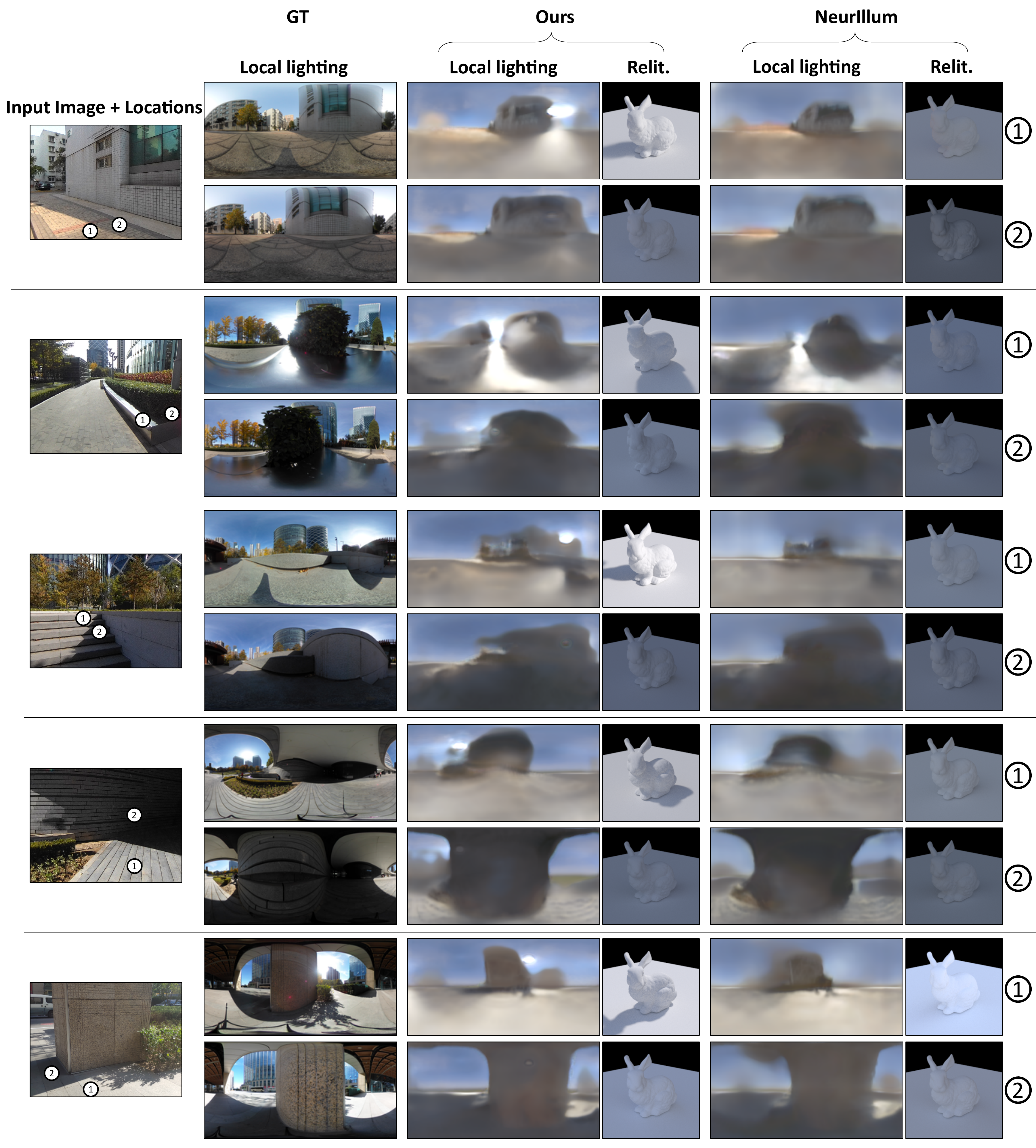}
     \caption{Spatially-varying lighting estimation and relighting results on real dataset.}
     \label{fig:relighting_results_local_real}
 \end{figure*}

\end{document}